\documentclass[journal]{IEEEtran}
\usepackage{graphicx}
\usepackage{amsmath}
\usepackage{amssymb}
\usepackage{booktabs}
\usepackage{subfloat}
\usepackage{multirow}
\usepackage{caption}
\usepackage{subcaption}
\usepackage{enumitem}
\usepackage{bbding}
\usepackage[T1]{fontenc}
\usepackage[ruled, linesnumbered]{algorithm2e}
\newcommand{\tabincell}[2]{\begin{tabular}{@{}#1@{}}#2\end{tabular}}%
\usepackage[pagebackref,breaklinks,colorlinks]{hyperref}

\begin{document}

\title{Semi-Supervised Domain Generalization with Evolving Intermediate Domain}

\author{Luojun~Lin, 
        Han~Xie, 
        Zhishu Sun,
        Weijie Chen,
        Wenxi~Liu$^{*}$,\\
        Yuanlong~Yu$^{*}$, 
        Lei~Zhang,~\IEEEmembership{Senior Member,~IEEE}
\thanks{L. Lin, H. Xie, Z. Sun, W. Liu and Y. Yu are with the College of Computer and Data Science, Fuzhou University, Fuzhou 350100, China (E-mails: linluojun2009@126.com; han\_xie@foxmail.com; siaimes@163.com; wenxi.liu@hotmail.com; yu.yuanlong@fzu.edu.cn).}
\thanks{W. Chen is with the Hikvision Research Institute, Hangzhou 310051, China (Email: chenweijie@hikvision.com)}
\thanks{L. Zhang is with the School of Microelectronics and Communication Engineering, Chongqing University, Chongqing 400044, China (Email: leizhang@cqu.edu.cn)}
\thanks{*Corresponding author: W.~Liu and Y.~Yu}
\thanks{Manuscript received April 27, 2023.}
}

\maketitle

\begin{abstract}
Domain Generalization (DG) aims to generalize a model trained on multiple source domains to an unseen target domain. The source domains always require precise annotations, which can be cumbersome or even infeasible to obtain in practice due to the vast amount of data involved. Web data, however, offers an opportunity to access large amounts of unlabeled data with rich style information, which can be leveraged to improve DG. From this perspective, we introduce a novel paradigm of DG, termed as \emph{Semi-Supervised Domain Generalization (SSDG)}, to explore how the labeled and unlabeled source domains can interact, and establish two settings, including the close-set and open-set SSDG. The close-set SSDG is based on existing public DG datasets, while the open-set SSDG, built on the newly-collected web-crawled datasets, presents a novel yet realistic challenge that pushes the limits of current technologies. 
A natural approach of SSDG is to transfer knowledge from labeled data to unlabeled data via pseudo labeling, and train the model on both labeled and pseudo-labeled data for generalization. Since there are conflicting goals between domain-oriented pseudo labeling and out-of-domain generalization, we develop a \emph{pseudo labeling phase} and a \emph{generalization phase} independently for SSDG. Unfortunately, due to the large domain gap, the pseudo labels provided in the pseudo labeling phase inevitably contain noise, which has negative affect on the subsequent generalization phase. Therefore, to improve the quality of pseudo labels and further enhance generalizability, we propose a cyclic learning framework to encourage a positive feedback between these two phases, utilizing an \emph{evolving intermediate domain} that bridges the labeled and unlabeled domains in a curriculum learning manner. Extensive experiments are conducted to validate the effectiveness of our method. It is worth highlighting that web-crawled data can promote domain generalization as demonstrated by the experimental results.
\end{abstract}

\IEEEpeerreviewmaketitle

\section{Introduction}
\IEEEPARstart{W}{ith} recent advances in deep learning, a major challenge has emerged: most deep models are trained under the \emph{i.i.d.} assumption that the training data and testing data are identically and independently distributed. However, the performance of such deep models always degrade drastically when facing real-world scenarios unsatisfied with \emph{i.i.d.} assumption. To address this issue, researchers concentrate on exploring \emph{Domain Generalization} (DG) techniques, with the goal of obtaining a robust model that can generalize to out-of-distribution data~\cite{li2020domain}.

Most DG studies follow the conventional supervised learning paradigm, where a deep model is trained on multiple source domains with ground-truth annotations~\cite{crossgrad, huang2020rsc, ddaig, mixstyle, cha2021swad}. The diverse styles from different source domains can enhance the deep model learn domain-invariant features and improve its generalization capacity~\cite{ddaig}. However, collecting sufficient annotated data with diverse styles from multiple domains can often be cumbersome or even infeasible in real-world scenarios. Fortunately, there is a way to overcome this challenge by using vast amounts of unlabeled data from the web, which is freely available and contains abundant style information so that can serve as auxiliary source data to enhance model generalization.

\begin{figure}[t]
\centering
\includegraphics[width=0.5\textwidth]{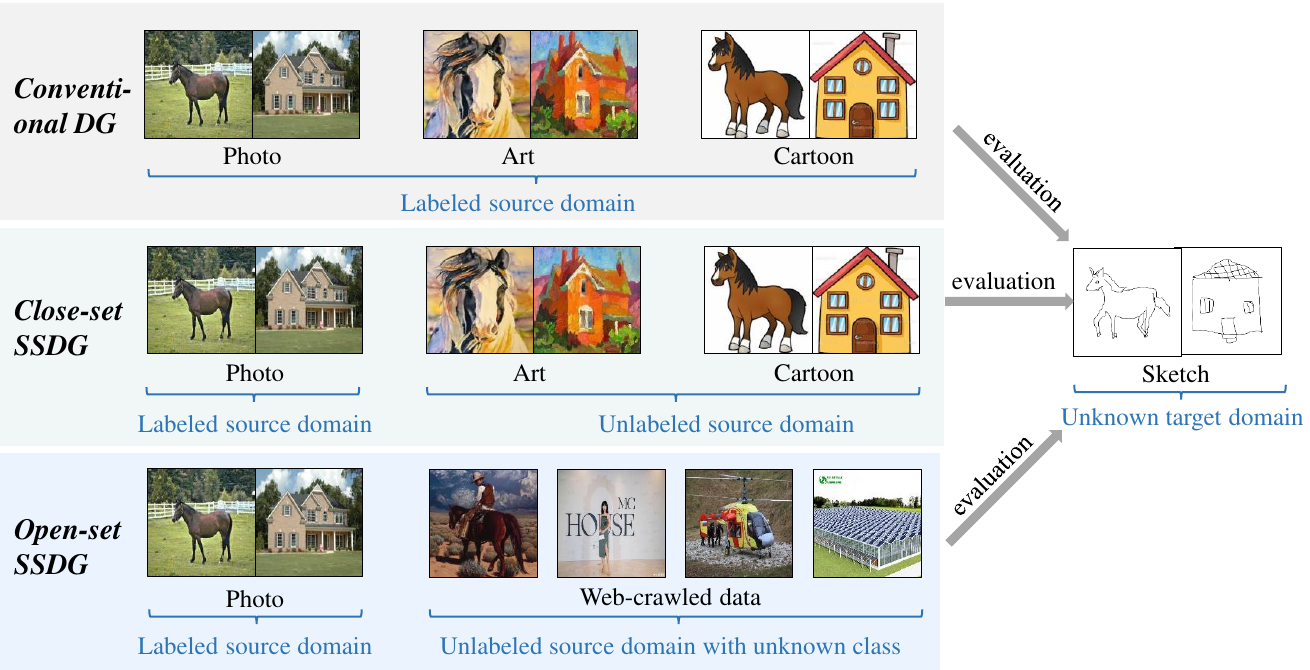}
\caption{Comparison between conventional domain generalization (DG) and semi-supervised domain generalization (SSDG). We propose two settings for SSDG, including the close-set SSDG and open-set SSDG.}
\vspace{-0.3cm}
\label{fig:ssdg}
\end{figure}

\begin{figure*}[ht]
\centering
\includegraphics[width=0.99\textwidth]{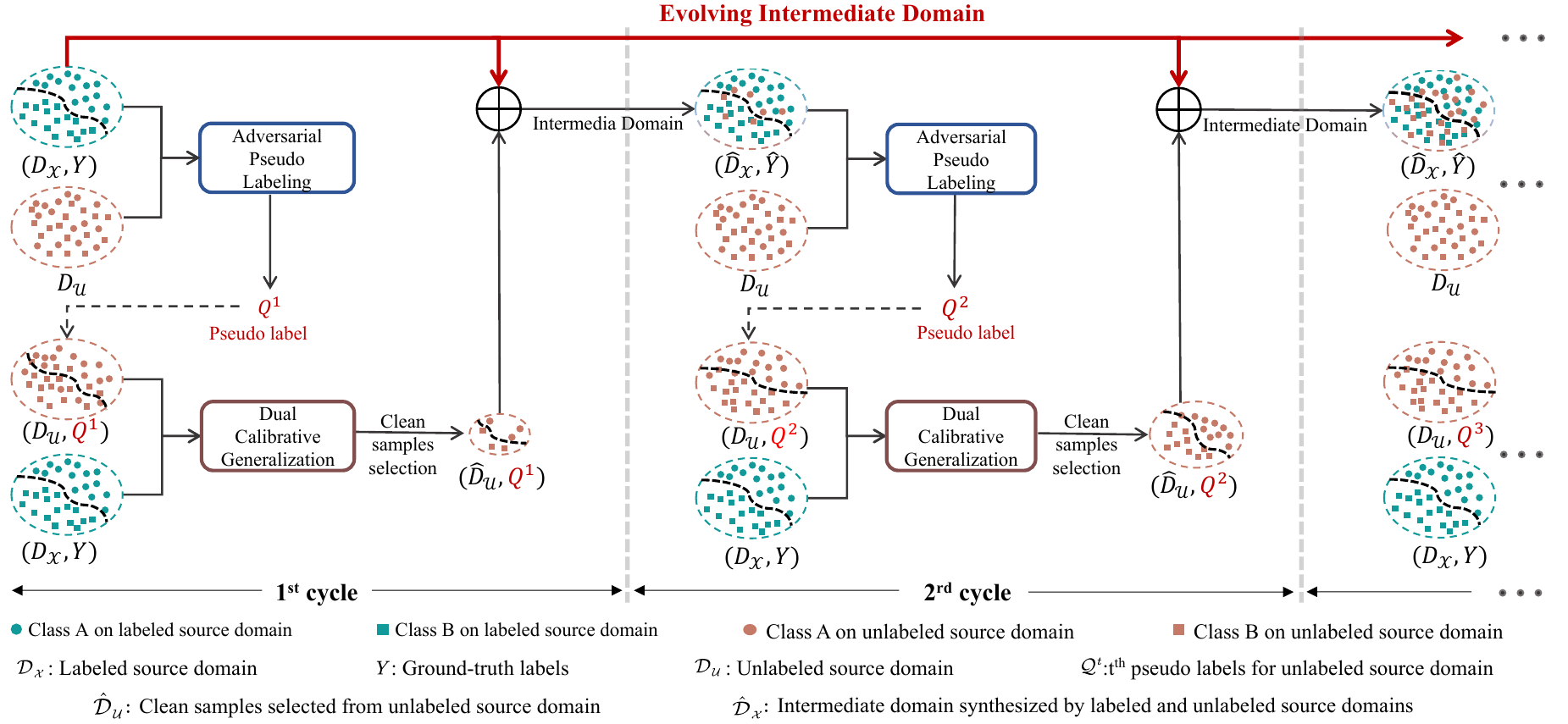}
\caption{The pipeline of our cycle framework for SSDG, which includes an \emph{Adversarial Pseudo Labeling} (APL) module for label propagation, and an \emph{Dual Calibrative Generalization} (DCG) module for domain generalization. APL and DCG modules are cyclically updated through the \emph{Evolving Intermediate Domain} (EID), which works to progressively reduce the domain gap within APL module for improving label propagation, thereby enhancing the generalization performance of DCG module.}
\label{fig:pipeline}
\end{figure*}

To this end, we propose a novel paradigm of DG, namely \emph{Semi-Supervised Domain Generalization (SSDG)}. Unlike the existing semi-supervised learning regime, the labeled and unlabeled training data in SSDG come from different data distributions. We also develop two different settings for SSDG, namely the \emph{close-set SSDG} and the \emph{open-set SSDG}, as depicted in Fig. \ref{fig:ssdg}. The close-set SSDG is established on existing public DG datasets, which are divided into homogeneous labeled and unlabeled source domains for training and the remaining domain as unseen target domain for evaluation. The open-set SSDG, on the other hand, more closely resembles practical scenarios, where the unlabeled source domains contain stylized web data that may include out-of-distribution (OOD) categories. The main challenge for these two settings is how to effectively utilize the unlabeled data. Compared to existing DG methods that have not attempted to deal with unlabeled data, our setting presents a novel yet realistic challenge and encourages pushing the limits of existing DG technologies.

A straightforward approach to SSDG is to disentangle the training period into two phases, a pseudo labeling phase, which propagates class information from labeled source domains to unlabeled source domains in order to produce high-quality pseudo labels, and a generalization phase, which trains a DG model using both the pseudo-labeled data and the true-labeled data. However, these pseudo labels can be unreliable and noisy due to the significant domain gap between labeled and unlabeled source domains. This can result in confusion for the DG model in decoupling label-related and domain-related features when it is trained to fit noisy labels, thereby degrading its generalization performance. In brief, the domain gap between the source domains has a direct impact on the quality of the pseudo labels, which in turn affects the generalization performance of the DG model.

To overcome the negative effect of the domain gap, in this study, we propose a cyclic learning framework that encourages positive interaction between the pseudo labeling phase and the generalization phase (see Fig.\ref{fig:pipeline}). In the pseudo labeling phase, we develop an \emph{Adversarial Pseudo Labeling} (APL) module that can flexibly incorporate current domain adaptation (DA) methods~\cite{dann, mcd}. The main difference from existing DA methods is that we update the labeled source domains using an \emph{Evolving Intermediate Domain} (EID) that bridges the labeled and unlabeled source domains in a curriculum learning manner. This design helps to gradually reduce the domain gap and therefore increasing the quality of pseudo labels, which subsequently, has positive impact on the training of DG model. 
However, despite the improvement in the pseudo labeling phase, there is still some stubborn noisy labels that can confuse the training of DG model in the generalization phase. To address this issue, we develop a \emph{Dual Calibrative Generalization} (DCG) module to filter the noisy labels using a dual network architecture in conjunction with style confusion training. In turn, the DCG module can provide positive feedback to the APL module by providing a clean set of pseudo-labeled data, interpolated with the true-labeled source domain to create an intermediate domain (EID) that can improve the APL for better pseudo-labeling. These processes are updated cyclically in our framework. As the cycle progresses, EID gets evolved, leading to a progressive enhancement in the APL module and a corresponding improvement in the DCG module.

Furthermore, we construct comprehensive benchmarks for both the close-set SSDG and open-set SSDG. For the close-set SSDG, we exploit existing DG datasets including Digits-DG~\cite{ddaig}, PACS~\cite{pacs}, and Office-Home~\cite{office-home}. For the open-set SSDG, we collect web-crawled data for PACS and Office-Home, respectively, and create two datasets including \emph{PACS-Webdata} and \emph{Office-Webdata}. 
We build extensive strong baselines by exploiting the domain adaptation, domain generalization, as well as semi-supervised learning methods on these SSDG benchmarks, which are always surpassed by our approach. Moreover, we perform thorough ablation studies to verify the contributions of each component in our framework. The extensive experiments presented in this paper provide a solid baseline in both close-set and open-set SSDG settings, which can benefit the future research in this field.

Overall, our contributions are summarized as follows:
\begin{itemize}[leftmargin=12pt, topsep=2pt, itemsep=0pt]
    \item We propose a novel paradigm of DG, termed as \emph{Semi-Supervised Domain Generalization (SSDG)}, which is more practical and significant than conventional DG that relies on fully-labeled source data.
    \item We develop two different settings for SSDG, including the close-set and open-set SSDG. Specifically, we construct two web-crawled datasets for the open-set SSDG, namely \emph{PACS-Webdata} and \emph{Office-Webdata}.
    \item We provide extensive strong baselines for SSDG varying among unsupervised domain adaptation, domain generalization, as well as semi-supervised learning methods.
    \item We propose a cyclic framework which includes an \emph{Adversarial Pseudo Labeling} (APL) module and a \emph{Dual Calibrative Generalization} (DCG) module. The two modules are mutually benefited via \emph{Evolving Intermediate Domain}, which progressively reduces domain discrepancy to enhance APL and then improve DCG for better generalization.
    \item Extensive experiments are conducted to validate the effectiveness of our method compared to strong baselines in both close-set and open-set SSDG scenarios. Our code and datasets have been already released: \url{https://github.com/MetaVisionLab/SSDG}
\end{itemize}

\section{Related Works}
\subsection{Domain Generalization (DG)} 
DG aims to train a model from multiple relevant style-variant source domains so as to generalize to the unseen target domain~\cite{li2018deep, sun2022dynamic}.
Early studies of DG focus on learning domain-invariant representation through domain adversarial training between domain classifier and feature extractor~\cite{li2018deep, matsuura2020domain}. 
Recent researches developed from meta learning fields simulate domain shift in training, with the purpose of enforcing model robust to domain shift in testing
~\cite{balaji2018metareg, li2019episodic,jia2022meta}.
Another line of solutions is based on the idea of data augmentation that aims to improve the style diversity in image-level~\cite{ddaig, crossgrad, xu2021fourier}, or feature-level~\cite{mixstyle, liuncertainty,zhongadversarial}.
Although current DG methods have shown promising outcomes, there are still some limitations, such as the heavy reliance on multiple labeled source domains with precise annotation, which can be overly cumbersome.

Compared with conventional DG, our SSDG is much closer to single DG where only a single labeled source domain is available in training. Most existing studies on single DG are based on adversarial domain augmentation that generates new training images to mimic virtual challenging domains in an adversarial way~\cite{huang2020rsc, qiao2020learning, l2d, fan2021adversarially}. For example, a recent work ~\cite{l2d} designs a style-complement module to synthesize images with diverse out-of-domain styles. 
Our work can serve as a complementary to existing single DG technologies by manually introducing unlabeled stylized images to enhance the performance of domain generalization.

\subsection{Domain Adaptation (DA)} 
DA is designed to mitigate the knowledge from labeled source domain to unlabeled target domain automatically, by training model with both source and target domain data. 
Previous DA methods tend to explore discrepancy-based methods to align distributions between source and target domains, by developing different distance metrics~\cite{MMD, wang2021rethinking}. Subsequently, researchers discover pre-defined distance functions cannot well describe the real distance between source and target domains, and adversarial training methods are developed to automatically align the features between source and target domains. For example, the feature extractor are adversarially trained with domain classifier to obtain domain-invariant features~\cite{GRL}, or adversarially trained with task-specific classifier to acquire task-oriented features~\cite{mcd, STAR,chen2022reusing, yang2022co}, or adversarially learn a balance between the feature transferability and feature discriminability~\cite{huang2022balancing}. Besides, generative adversarial networks are also utilized in DA field for domain mapping, \emph{e.g.}, mapping source images to target styles~\cite{CoGAN}, or in reverse~\cite{PixelDA, CYCADA}. Recently, some pseudo labeling methods are developed for DA through self-training~\cite{ chen2022self}.
In our method, all existing DA methods can be implemented in the APL module to produce pseudo labels, and EID can further improve the performance of these DA methods by gradually annealing domain discrepancy.

\subsection{Semi-Supervised Learning (SSL)} 
Semi-supervised learning (SSL) aims to improve model performance by incorporating knowledge from unlabeled data. One popular SSL technique is pseudo labeling, which involves assigning confident prediction classes as pseudo labels~\cite{pseudo-label}, where entropy minimization is usually used to ensure the pseudo labels being confident enough~\cite{grandvalet2005semi}. Another well-known SSL method is consistency regularization, which assumes that a model response should remain consistent after changes to the input or the model itself. These changes could include data augmentation~\cite{MixMatch}, time ensembling of model in different time steps~\cite{Temporal}, an exponential moving average of the model parameters in different time steps~\cite{Mean-teacher}, or adversarial perturbations to the model parameters~\cite{VAT}. There are also some hybrid works that combine pseudo labeling and consistency regularization techniques, such as MixMatch~\cite{MixMatch}, FeatMatch~\cite{kuo2020featmatch} and FixMatch~\cite{FixMatch}. In this work, Self-Supervised Domain Generalization (SSDG) can be considered a special case of SSL where the labeled and unlabeled data come from different distributions and the goal is to generalize the model to out-of-distribution data.

\subsection{Noisy Label Learning (NLL)} 
When training with noisy labels, a \emph{memorization effect} is discovered in deep networks that the model is prone to fit easy (clean) samples and then gradually overfit hard (noisy) samples~\cite{memorization, song2022learning}. It indicates that samples with small-loss are likely to be with clean labels, which suggests to update networks on such samples to avoid overfitting on noisy labels. Hence, researchers are inspired to design effective criterion to select small-loss (clean) samples, \emph{e.g.}, MentorNet~\cite{mentornet} using mentor-net to provide sample selection curriculum for student-net.
Another line of works use a small batch of trusted data to recover and refine label distribution from corrupted labels~\cite{xu2022trusted}.
Co-teaching~\cite{Co-teaching} trains two networks mutually that each network is updated on the clean samples selected by its peer network per feed-forward. Based on such inspirations, our DCG module is designed to filter the noisy labels from APL module. Moreover, we focus not only on filtering noisy labels, but also on how to improve the generalization ability of DCG module.

\section{Methodology}
\subsection{Overview} \label{sec:overview}
We only focus on $K$-way classification problem of SSDG in this paper. For simplicity, we only consider the situation that the labeled data is drawn from a single source domain $\mathcal{D}_\mathcal{X}=\{(x_i, y_i)\}_{i=1}^{n_l}$, and the unlabeled data $\mathcal{D}_\mathcal{U}=\{u_i\}_{i=1}^{n_u}$ is drawn from multiple source domains. As shown in Fig. \ref{fig:pipeline}, we treat SSDG as a progressive evolving two-phase task, including \emph{Adversarial Pseudo Labeling} (APL) and \emph{Dual Calibrative Generalization} (DCG), which are mutually benefited via an \emph{Evolving Intermediate Domain} (EID) to bridge the labeled and unlabeled data progressively.

Before introducing our framework, we have to emphasize that the motivations of APL and DCG are totally different, where the former one aims to learn accurate pseudo label for the unlabeled data, while the latter one aims to generalize well on agnostic target domain. Actually, APL prefers a smaller domain gap so as to propagate the label information from the labeled data to the unlabeled data, while DCG prefers a larger domain gap so as to learn domain-invariant features for better generalization. Nonetheless, there still exist connections between these two phases. First, the performance of DCG heavily relies on the accuracy of the pseudo labels provided by APL, because the noisy pseudo labels will confuse DCG to disentangle label-related features and domain-related features. Second, the performance of APL heavily relies on the domain gap between the labeled and unlabeled data, which means we can improve the accuracy of the pseudo labels on the unlabeled data by narrowing the domain gap. Inspired by the aforementioned considerations, we develop EID to connect these two phases by promoting a cycle learning framework. Our method can be summarized as follows: 
\begin{enumerate}
    \item APL is designed to generate pseudo label for unlabeled data which can be implemented by any existing unsupervised domain adaptation method; 
    \item Through exploiting the pseudo labels, DCG performs style confusion training and label diversity regularization for domain generalization. Besides, DCG can be further used to filter the noisy pseudo labels provided by APL module, when finishing training on each cycle; 
    \item EID builds an intermediate domain via fusing the labeled data and the clean set of the unlabeled data. Replacing the labeled data with the intermediate domain in APL, APL and DCG are promoted to drive next cycle optimization in a curriculum way, in order to benefit each other mutually. 
\end{enumerate}

\subsection{Progressive Adversarial Pseudo Labeling} \label{sec:pseudo_label}
As shown in Fig. \ref{fig:pipeline}, our adversarial pseudo labeling (APL) module is designed to generate pseudo labels $Q=\{q_i\}_{i=1}^{n_u}$ for the unlabeled data $\mathcal{D}_\mathcal{U}$ driven by the evolving intermediate domain $\widehat{\mathcal{D}}_\mathcal{X}$: 
\begin{equation}
\begin{aligned}
    & Q=\mathop{\arg\max}p_{\phi}(\mathcal{D}_\mathcal{U}), \\
    & s.t. \quad \phi = \mathop{\arg\min}_\phi \mathcal{L}_{apl}(\phi; \widehat{\mathcal{D}}_\mathcal{X}, \mathcal{D}_\mathcal{U}),
    \label{eq:APL}
\end{aligned}
\end{equation}
where $p_{\phi}$ is the class probability predicted by the network with weights of $\phi$. The intermediate domain $\widehat{\mathcal{D}}_\mathcal{X}$ will be introduced in Sec.\ref{sec:evolving}. In the first cycle period, we initialize $\widehat{\mathcal{D}}_\mathcal{X}$ with the labeled source domain ${\mathcal{D}}_\mathcal{X}$. In Eq.\ref{eq:APL}, we aim to propagate the label information from $\widehat{\mathcal{D}}_\mathcal{X}$ to $\mathcal{D}_\mathcal{U}$ by employing the supervision signals $\mathcal{L}_{apl}$ from existing DA methods, such as MCD~\cite{mcd}, which performs adversarial training between feature extractor and task-specific classifiers. Note that it can be replaced by any other DA methods. As the training goes on, $\widehat{\mathcal{D}}_\mathcal{X}$ is kept evolved to anneal domain gap between labeled and unlabeled domains, which advances the quality of pseudo label to promote the optimization of DCG.

\subsection{Dual Calibrative Generalization}
\begin{figure}[t]
\centering
\footnotesize
\includegraphics[width=0.48\textwidth]{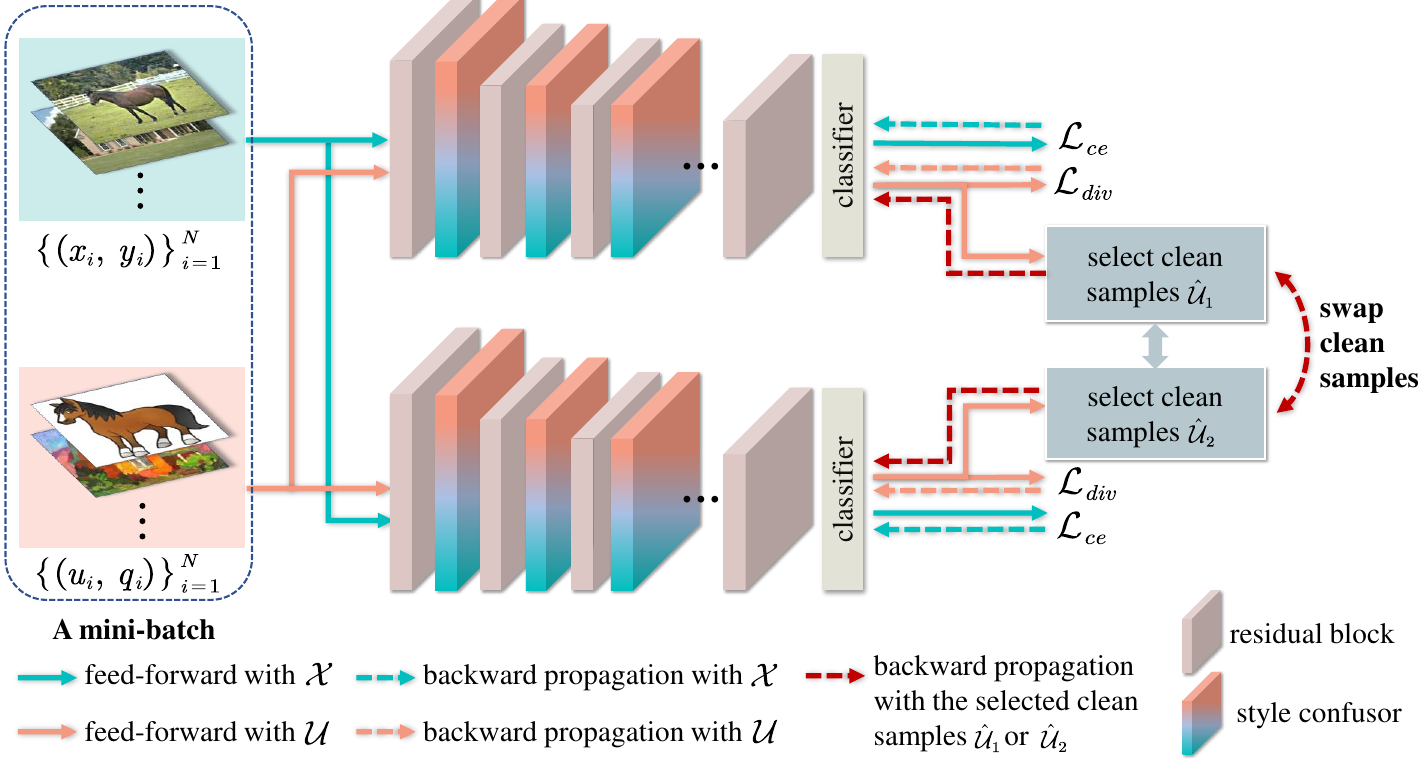}
\caption{Dual calibrative generalization (DCG) with style confusion training and label diversity regularization.}
\vspace{-0.1cm}
\label{fig:dcg}
\end{figure}

The pseudo labels provided by APL are inevitable to be noisy, which will confuse the model to disentangle label-related features and domain-related features. From this perspective, we construct DCG to calibrate pseudo labels when learning domain generalization, which is equipped with style confusion training and label diversity regularization jointly. 

As shown in Fig.\ref{fig:dcg}, DCG consists of two identical subnetworks, parameterized with $\theta_1$ and $\theta_2$, respectively. Each subnetwork is calibrated by its peer network through updating the clean samples selected by each other. Specifically, each subnetwork is initialized randomly and fed with the same mini-batch that contains $N$ labeled samples $\mathcal{X}=\{(x_1, y_1), ..., (x_N, y_N)\}$ drawn from $\mathcal{D}_\mathcal{X}$, and $N$ pseudo-labeled samples $\mathcal{U}=\{(u_1, q_1), ..., (u_N, q_N)\}$ drawn from $\mathcal{D}_\mathcal{U}$. 
In this way, the training objective of DCG in each iteration is composed of a labeled source part $\mathcal{X} (\mathcal{X} \subseteq  \mathcal{D}_\mathcal{X})$ and an unlabeled source part $\mathcal{U} (\mathcal{U} \subseteq  \mathcal{D}_\mathcal{U})$. The classification loss for the labeled source part is defined as:
\begin{equation}
    \min\limits_{\theta_1, \theta_2} \mathcal{L}_{ce}(\theta_1; \mathcal{X}) + \mathcal{L}_{ce}(\theta_2; \mathcal{X}).
\label{eq:labeled_loss}
\end{equation}
It can regularize the optimization of the unlabeled source part. With the guidance of $\mathcal{X}$, the model is able to select the clean set of $\mathcal{U}$ based on a small-loss strategy. It suggests that the samples with small classification loss tend to be clean and then can be used to update the model. In addition, we reduce the selection rate of small-loss samples as the training goes on, since the \emph{memorization effect} reveals that models tend to learn easy (clean) samples before over-fitting to hard (noisy) samples~\cite{memorization}. 
Based on this sampling strategy, we can obtain two distinct clean sets, namely $\widehat{\mathcal{U}}_1$ and $\widehat{\mathcal{U}}_2$ ($\widehat{\mathcal{U}}_1, \widehat{\mathcal{U}}_2 \subseteq \mathcal{U}$), selected by the subnetworks $\theta_1$ and $\theta_2$, respectively. The two clean sets are swapped to update subnetworks, so that the classification loss of the unlabeled source part is defined as:
\begin{equation}
    \min\limits_{\theta_1, \theta_2} \mathcal{L}_{ce}(\theta_1; \widehat{\mathcal{U}}_2) + \mathcal{L}_{ce}(\theta_2; \widehat{\mathcal{U}}_1).
\label{eq:clean_loss}
\end{equation}
In this way, the two subnetworks will eventually have different decision boundaries, which will increase their noise tolerance to avoid overfitting to the noisy pseudo labels.

\subsubsection{Style Confusion Training}
To improve the generalization capability of DCG, we employ a low-cost style confusion training method to insert \emph{style confusor} into each subnetwork to enrich the styles in feature level. Developed from adaptive instance normalization (AdaIN)~\cite{nam2018bnDG}, the style confusor is designed to confuse the styles between the labeled and unlabeled data in feature space. Specifically, the styles refer to the statistics of instance normalization (IN)~\cite{ulyanov2016IN}. The statistics includes the mean $\mu=\{\mu_x^1, ..., \mu_x^N,\mu_u^1, ..., \mu_u^N\}$ and the variance $\sigma=\{\sigma_x^1, ..., \sigma_x^N,\sigma_u^1, ..., \sigma_u^N\}$ calculated from each input mini-batch $\{x_1, ..., x_N,u_1, ..., u_N\}$. The statistics are randomly shuffled in the batch dimension to obtain a new mean $\tilde{\mu}$ and new variance $\tilde{\sigma}$, which can be considered a new feature style. Via a linear combination between $\{\mu, \sigma\}$ and $\{\tilde{\mu}, \tilde{\sigma}\}$, we can achieve the transformation parameters $\{\beta,\gamma\}$:
\begin{equation}
    \beta = \lambda\mu + (1 - \lambda) \tilde{\mu}, \quad
    \gamma = \lambda\sigma + (1 - \lambda)\tilde{\sigma},
\label{eq:transformation_params}
\end{equation}
where $\lambda$ is randomly sampled from a Beta distribution $B(0.5, 0.5)$. We utilize the transformation parameters $\{\beta,\gamma\}$ to transfer the original features into new style space:
\begin{equation}
    \widehat{F} = \gamma \cdot \frac{F - \mu}{\sigma + \epsilon} + \beta,
\label{eq:transform}
\end{equation}
where $\epsilon$ is a fixed constant with negligible value. $F=\{F_x^1,...,F_x^N, F_u^1,...,F_u^N\}$ and $\widehat{F}=\{\widehat{F}_x^1,...,\widehat{F}_x^N,\widehat{F}_u^1,...,\widehat{F}_u^N\}$ represent the features in a mini-batch before and after transformation, respectively. Note that the style confusor is omitted during inference stage.

\subsubsection{Label Diversity Regularization}
Eq.\ref{eq:clean_loss} indicates that only the clean part of the pseudo-labeled data is reserved to optimize the model, leading to inefficient utilization of the remaining unlabeled data. To remedy this issue, we employ a label diversity regularization on the whole unlabeled data part $\mathcal{U}$ in each mini-batch. It is designed to minimize the instance entropy for instance discrimination and maximize the global entropy for label diversification, as defined by:
\begin{equation}
\begin{aligned}
    \mathcal{L}_{div}(\theta; \mathcal{U}) =  &\sum \mathbb{E}_{u_i\in\mathcal{U}} p_{\theta}(u_i) \log \mathbb{E}_{u_i\in\mathcal{U}} p_{\theta}(u_i) \\
    & - \mathbb{E}_{u_i\in\mathcal{U}} \sum p_{\theta}(u_i) \log p_{\theta}(u_i),
\end{aligned}
\label{eq:divloss}
\end{equation}
where $p_{\theta}$ denotes $K$-way prediction probability of the subnetwork parameterized with $\theta$ ($\theta$ represents $\theta_1$ or $\theta_2$ here). 
The first term expects for prediction diversification based on class balance assumption, while the second term requires instance discrimination based on entropy minimization theory. According to the aforementioned discussion, the final objective of DCG with input mini-batch $\{\mathcal{X}, \mathcal{U}\}$ is summarized as:
\begin{equation}
\begin{split}
    \min\limits_{\theta_1, \theta_2}  & \mathcal{L}_{ce}(\theta_1; \mathcal{X}) + \mathcal{L}_{ce}(\theta_2; \mathcal{X}) + \mathcal{L}_{ce}(\theta_1; \widehat{\mathcal{U}}_2) + \\
    &\mathcal{L}_{ce}(\theta_2; \widehat{\mathcal{U}}_1)  + \mathcal{L}_{div}(\theta_1; \mathcal{U}) + \mathcal{L}_{div}(\theta_2; \mathcal{U}) .
\label{eq:allloss}
\end{split}
\end{equation}
\begin{figure}[t]
\centering
\footnotesize
\includegraphics[width=0.45\textwidth]{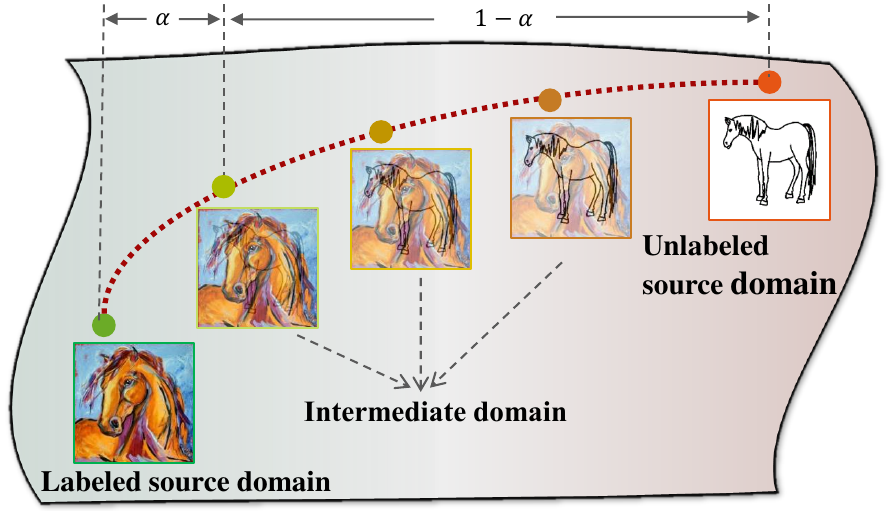}
\vspace{-0.1cm}
\caption{Illustration of intermediate domain.}
\vspace{-0.1cm}
\label{fig:eid}
\end{figure}

\begin{algorithm*}[t]
\caption{Training Procedure of Our Method}
\label{algorithm}
\SetAlgoLined
\LinesNumbered
\KwIn{Labeled source domain $\mathcal{D}_\mathcal{X}$; unlabeled source domain $\mathcal{D}_\mathcal{U}$; weights $\phi$, learning rate $\eta_{\phi}$, training epochs $T_{apl}$ for APL module; weights $\{\theta_1, \theta_2$\},  learning rate $\eta_{\theta}$, training epochs $T_{dcg}$ for DCG modules;
maximum iterations $N_{max}$ per epoch for DCG; 
clean rate $r(t)$ with hyper-parameters $\delta$ and $T_{k}$ for DCG; 
clean rate $\mathcal{R}$ and maximum cycles $\mathcal{C}$ for EID.
}
\KwOut{$\theta_1$ and $\theta_2$.}
Initialize intermediate domain ${\widehat{\mathcal{D}}_\mathcal{X} = \mathcal{D}_\mathcal{X}}$\;
\For{$c = 1$ to $\mathcal{C}$}{
    \For{$t = 1$ to $T_{apl}$}{
    Update $\phi = \phi - \eta_{\phi} \cdot \mathcal{r} \mathcal{L}_{apl}(\phi; \widehat{\mathcal{D}}_\mathcal{X}, \mathcal{D}_\mathcal{U})$; \tcp*[f]{update APL}\; 
    }
    Generate pseudo labels by Eq. \ref{eq:APL}\;
    Initialize clean rate $r(1)=1$\;
    \For{$t = 1$ to $T_{dcg}$}{
        \For {$n = 1$ to $N_{max}$}{
            Fetch a mini-batch $\mathcal{B}= \{\mathcal{X}, \mathcal{U} | \mathcal{X} \subseteq \mathcal{D}_\mathcal{X}, \mathcal{U} \subseteq \mathcal{D}_\mathcal{U} \} $\;
            Acquire $\widehat{\mathcal{U}}_1 = \mathop{\arg\min}_{\mathcal{U}':|\mathcal{U}'|\geqslant r(t)|\mathcal{U}|} \mathcal{L}_{ce}(\theta_1; \mathcal{U}')$;
            \tcp*[f]{small-loss sampling}\;
            Acquire $\widehat{\mathcal{U}}_2 = \mathop{\arg\min}_{\mathcal{U}':|\mathcal{U}'|\geqslant r(t)|\mathcal{U}|} \mathcal{L}_{ce}(\theta_2; \mathcal{U}')$;
            \tcp*[f]{small-loss sampling}\;
            Update $\theta_1 = \theta_1 - \eta_{\theta} \cdot 
            [\mathcal{r}\mathcal{L}_{ce}(\theta_1; \mathcal{X}) +  
            \mathcal{r}\mathcal{L}_{ce}(\theta_1; \widehat{\mathcal{U}}_2) + 
           \mathcal{r} \mathcal{L}_{div}(\theta_1; \mathcal{U})]$; \tcp*[f]{update DCG}\; 
            Update $\theta_2 = \theta_2 - \eta_{\theta} \cdot 
            [\mathcal{r}\mathcal{L}_{ce}(\theta_2; \mathcal{X}) +  
            \mathcal{r}\mathcal{L}_{ce}(\theta_2; \widehat{\mathcal{U}}_1) + 
            \mathcal{r} \mathcal{L}_{div}(\theta_2; \mathcal{U})]$; \tcp*[f]{update DCG}\; 
        }
        Update clean rate $r(t+1)=1-\min(\frac{t}{T_{k}}\delta, \delta)$; \tcp*[f]{update clean rate for DCG}\;
    }
    Acquire candidate clean set $\bar{\mathcal{D}}_\mathcal{U}$ by Eq. \ref{eq:agreement}; \tcp*[f]{agreement-based sampling} \;
    Acquire final clean set $\widehat{\mathcal{D}}_\mathcal{U}$ by Eq. \ref{eq:small_loss}; \tcp*[f]{small-loss sampling}\;
    Update intermediate domain $\widehat{\mathcal{D}}_\mathcal{X}$ by Eq. \ref{eq:mixup}; \tcp*[f]{update intermediate domain}\;
}
\end{algorithm*}

\subsection{Evolving Intermediate Domain} \label{sec:evolving}
As discussed above, the more accurate pseudo labels can promote DCG to disentangle domain-related feature and label-related features and thus improving its generalization capability. However, the existence of domain gap between labeled source domain $\mathcal{D}_\mathcal{X}$ and unlabeled source domain $\mathcal{D}_\mathcal{U}$ limits APL to provide high-quality pseudo labels. In order to reduce the domain gap, we propose an \emph{Evolving Intermediate Domain} (EID) to synthesize intermediate domain $\widehat{\mathcal{D}}_\mathcal{X}$ to bridge the labeled and unlabeled source domain data. Actually, the intermediate domain $\widehat{\mathcal{D}}_\mathcal{X}$ is the interpolation of the labeled data and the clean set of unlabeled data, as shown in Fig. \ref{fig:eid}.

The intermediate domain $\widehat{\mathcal{D}}_\mathcal{X}$ will serve as the new source data for the upcoming training cycle of APL, emphasizing the importance of maintaining the accuracy of the interpolated labels in the intermediate domain. And this accuracy is reflected in the quality of the clean set provided by DCG. To achieve this, we adopt an agreement-based sampling strategy to produce a candidate clean set $\bar{\mathcal{D}}_\mathcal{U}$, based on the predictions of DCG on unlabeled data $\mathcal{D}_\mathcal{U}$. This requires that only when the two subnetworks agree on the prediction of sample $u_i$, will the sample and its corresponding prediction $q_i$ be included in the candidate clean set $\bar{\mathcal{D}}_\mathcal{U}$, as defined by:
\begin{equation}
\begin{aligned}
   & \bar{\mathcal{D}}_\mathcal{U}=\bar{\mathcal{D}}_\mathcal{U}\cup(u_i, q_i),  \\
   & s.t.  
    \mathop{\arg\max} p_{\theta_1}(u_i)= \mathop{\arg\max} p_{\theta_2}(u_i),\\ & \and q_i=\mathop{\arg\max} p_{\theta_1}(u_i),
\label{eq:agreement}
\end{aligned}
\end{equation}
where $p_{\theta_1}(u_i)$ and $p_{\theta_2}(u_i)$ denote $K$-way prediction probabilities from two subnetworks on the $i$-th unlabeled sample. Further, we adopt small-loss sampling strategy on $\bar{\mathcal{D}}_\mathcal{U}$ with a clean rate of $\mathcal{R}$ to obtain a higher-quality clean set:
\begin{equation}
    \widehat{\mathcal{D}}_\mathcal{U} = \mathop{\arg\min}\limits_
    {\mathcal{D}':|\mathcal{D}'|\geq \mathcal{R}| \bar{\mathcal{D}}_\mathcal{U}|} \mathcal{L}_{ce}(\theta_1; \mathcal{D}')+\mathcal{L}_{ce}(\theta_2; \mathcal{D}').
\label{eq:small_loss}
\end{equation}

Then, the clean set $\widehat{\mathcal{D}}_\mathcal{U}$ is linearly mixed with labeled data $\mathcal{D}_\mathcal{X}$ to synthesize the intermediate domain $\widehat{\mathcal{D}}_\mathcal{X}$:
\begin{equation}
\begin{split}
    &{\hat{x}_i} = \alpha x_i + (1 - \alpha) u_j,  \\
    &{\hat{y}_i} = E(\alpha) y_i + (1 - E(\alpha)) q_j,  
\label{eq:mixup}
\end{split}
\end{equation}
where $(x_i,y_i) \in \mathcal{D}_\mathcal{X}\,\,$, $ (u_j,q_j) \in \widehat{\mathcal{D}}_\mathcal{U}\,\,$, $({\hat{x}_i}, {\hat{y}_i}) \in \widehat{\mathcal{D}}_\mathcal{X}$ and $\alpha$ is a weighting map with the same spatial resolution as the input data to control the mixing mechanism. $E(\alpha)$ denotes a scalar value, as well as the average of $\alpha$. For simplicity, we only discuss three mixing mechanisms in this paper, including CutMix~\cite{Cutmix}, MixUp~\cite{Mixup}, and our self-defined ``X+U'', which can be extended to more mixing manners in future works. 
(1) For CutMix, $\alpha$ 
is filled with 1 pasted by a segment of 0 randomly. 
(2) For MixUp, $\alpha$ shares the same value that is in the range of $[0,1]$ and randomly drawn from a Beta distribution.
(3) For ``X+U'', $\alpha$ shares the same value of 0 or 1, which means we directly combine $\mathcal{D}_\mathcal{X}$ and $\widehat{\mathcal{D}}_\mathcal{U}$ to form the intermediate domain
$\widehat{\mathcal{D}}_\mathcal{X}$.

Compared to the original source data $\mathcal{D}_\mathcal{X}$, the new source data, \emph{i.e.}, the intermediate domain $\widehat{\mathcal{D}}_\mathcal{X}$, exhibits a smaller domain discrepancy with the unlabeled data $\mathcal{D}_\mathcal{U}$. This reduction in domain discrepancy is favorable for enhancing APL to provide high-quality pseudo labels, which in turn enables DCG to improve its generalization performance. As the training cycles progress, the clean set of unlabeled data provided by DCG is increased and diversified, and the intermediate domain is gradually evolved, leading to a continuous mutually beneficial relationship between APL and DCG.
The training procedure is illustrated in Algorithm \ref{algorithm}.

\section{Constructing Open-Set SSDG Benchmark}
In the real world, obtaining web-crawled data is easy and low-cost, making it a popular choice for applying to DG due to its vast diversity of styles. To support this viewpoint, we construct open-set SSDG benchmarks by collecting two web-crawled dataset called \emph{PACS-Webdata} and \emph{Office-Webdata} to serve as the auxiliary source data for PACS~\cite{pacs} and Office-Home~\cite{office-home}, respectively. More detailed settings of open-set SSDG can refer to Section \ref{sec:exp-open-set-ssdg}.

\begin{figure}[ht]
\centering
     \begin{subfigure}[b]{1.0\linewidth}
         \centering
         \includegraphics[width=0.9\linewidth]{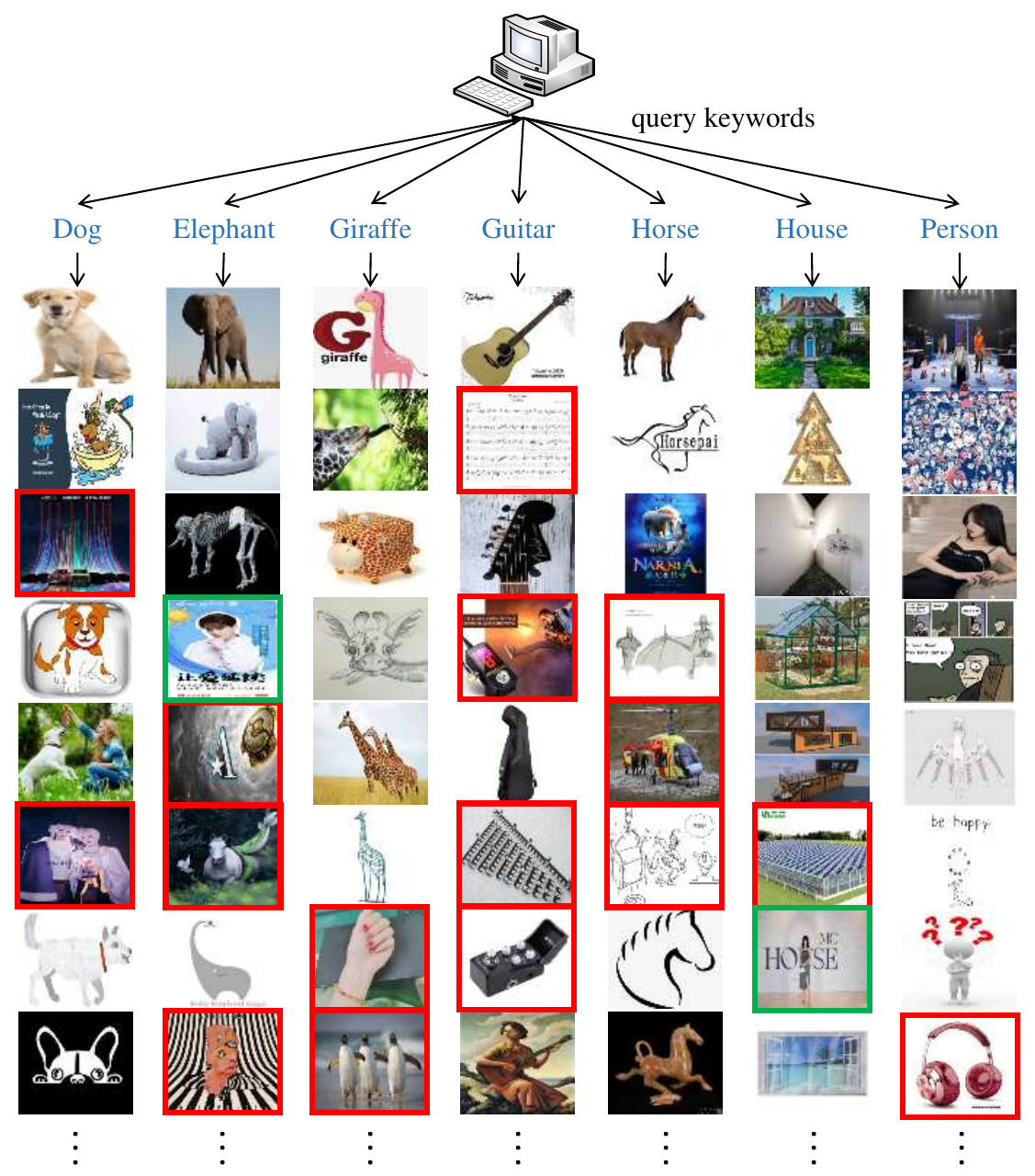}
          \vspace{-0.2cm}
         \caption{PACS-Webdata}
     \end{subfigure}
     \begin{subfigure}[b]{1.0\linewidth}
         \centering
         \includegraphics[width=0.9\linewidth]{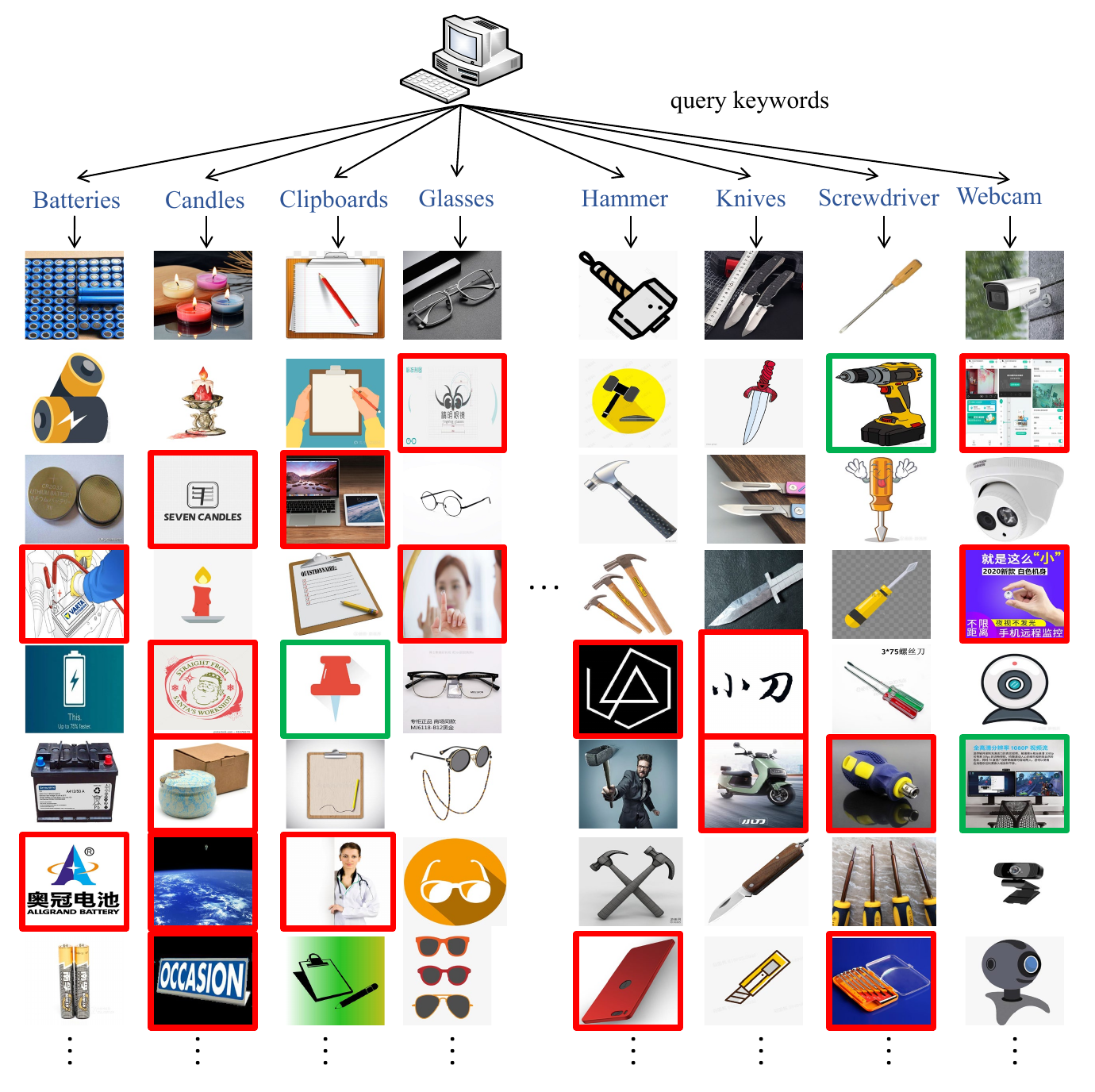}
         \vspace{-0.2cm}
         \caption{Office-Webdata}
     \end{subfigure}
\vspace{-0.5cm}
\caption{Examples from PACS-Webdata and Office-Webdata, where the noisy samples are indicated by either a red box or a green box. The red box signifies that the sample belongs to an out-of-distribution (OOD) class, and the green box indicates that the sample belongs to an in-distribution class but the web-label assigned to it is incorrect.}
\vspace{-0.4cm}
\label{fig:webdata}
\end{figure}

For PACS-Webdata, we collect over $8,000$ web-crawled images by using class names as query keywords for a search engine and crawling a similar number of images for each class. 
For Office-Webdata, we collect over $360,000$ web-crawled images, with each class containing $400 \sim 500$ images. 
Fig. \ref{fig:webdata} displays several search results, revealing that the web-crawled data is distinct from existing DG datasets as it contains images with rich styles and diverse backgrounds. We also found that the query keywords can serve as approximate class labels for the web-crawled images, referred to as ``web-labels''. However, these web-labels come with unavoidable noise. As shown in Fig. \ref{fig:webdata}, there exists two types of noise in web-labels: 1) the true labels of images belong to in-distribution classes but are inconsistent with the query keyword and 2) the true labels of images belong to out-of-distribution (OOD) classes, also known as OOD. Table \ref{tab:pacs-webdata} presents the OOD ratio per category, indicating that OOD occupies a large proportion in such web-crawled data,  close to one-sixth of the total number of images. This highlights that OOD is the primary difference between the close-set and open-set SSDG. Moreover, OOD is also the main challenge in using web-crawled data to improve DG performance. Further analysis is discussed in the experiments.
\begin{table}[t]  
\centering
\footnotesize
\setlength\tabcolsep{2pt}
{
\begin{tabular}{c|c|c|c|c|c|c|c|c}  
\toprule  
Amount & Dog & Elephant & Giraffe & Guitar & Horse & House & Person & Avg.\cr
\midrule
Total number & 742 & 597 & 1001 & 952 & 699 & 590 & 662 & 5243   \cr
OOD number & 108 & 104 & 161 & 145 & 111 & 86 & 124 & 839\cr
OOD rate (\%) & 14.56 & 17.42 & 16.08 & 15.23 & 15.88 & 14.58 & 18.73 & 16.00 \cr
\bottomrule  
\end{tabular}}
\vspace{-0.1cm}
\caption{The OOD statistics of PACS-Webdata.}
\vspace{-0.4cm}
\label{tab:pacs-webdata}
\end{table}

\section{Experiments}
\begin{table*}[ht!]  
\centering
\footnotesize
\resizebox{1.0\textwidth}{!}{
\begin{tabular}{l|l|ccc|ccc|ccc|ccc|c}  
\toprule  
Backbone\!\! & PL Method & P$\scriptstyle{\rightarrow}$A & P$\scriptstyle{\rightarrow}$C & P$\scriptstyle{\rightarrow}$S & 
 A$\scriptstyle{\rightarrow}$P & A$\scriptstyle{\rightarrow}$C & A$\scriptstyle{\rightarrow}$S & 
 C$\scriptstyle{\rightarrow}$P & C$\scriptstyle{\rightarrow}$A & C$\scriptstyle{\rightarrow}$S & 
 S$\scriptstyle{\rightarrow}$P & S$\scriptstyle{\rightarrow}$A & S$\scriptstyle{\rightarrow}$C &
 Avg.\\
\midrule
\multirow{4}{*}{ResNet-18.} & Ours w/ DANN~\cite{dann}   & 61.32 & 41.16 & 26.45 & \textbf{95.39} & 70.01 & 65.30 & \textbf{86.42}  & 77.10  & 70.62 & 60.89 & 61.54 & 65.33 & 65.13    \\
& Ours w/ MME~\cite{MME}  & \textbf{75.94}  & 68.11 & 63.50 & 94.94 & 68.37 & \textbf{68.46} & 84.94 & 76.40 & 71.47 & 64.48 & \textbf{67.33} & 69.46 & 72.78 \\
& Ours w/ MCD~\cite{mcd}  & 75.52 & \textbf{70.98} & \textbf{64.01}  & 94.94  & \textbf{71.82} & 67.18 & 84.64  & \textbf{77.39} & \textbf{72.24} & \textbf{67.21} & 66.86 & \textbf{72.83} & \textbf{73.80}   \\
\midrule
ResNet-18. & \multirow{4}{*}{Ours w/ MCD~\cite{mcd}} & 75.52 & 70.98 & 64.01  & 94.94  & 71.82 & 67.18 & 84.64  & 77.39 & 72.24 & 67.21 & 66.86 & 72.83 & 73.80   \\
ResNet-50. & & 71.84 & 70.65 & 66.06  & \textbf{95.17}  & 74.34 & \textbf{67.25} & \textbf{90.43}  & \textbf{79.81} & 69.62 & \textbf{73.57} & 66.10 & 74.44 & 74.94 \\
ResNet-101. & & \textbf{75.69} & \textbf{72.55} & \textbf{68.33}  & 94.43  & \textbf{77.95} & 62.88 & 86.84  & 79.54 & \textbf{73.10} & 70.32 & \textbf{76.61} & \textbf{76.52} & \textbf{76.23} \\
\bottomrule  
\end{tabular}}
\caption{Ablation studies in terms of APL module with different pseudo labeling (PL) methods or different backbones on \textbf{PACS} dataset. ``P$\rightarrow$A'' means P is the labeled source domain, A is the testing target domain, and the remaining domains are treated as the unlabeled source domains. Same representation in the following tables.}
\label{tab:ablation_apl}
\end{table*}
\begin{table*}[ht!]  
\centering
\footnotesize
\resizebox{1.0\textwidth}{!}{
\begin{tabular}{l|ccc|ccc|ccc|ccc|c}  
\toprule  
 Method & P$\scriptstyle{\rightarrow}$A & P$\scriptstyle{\rightarrow}$C & P$\scriptstyle{\rightarrow}$S & 
 A$\scriptstyle{\rightarrow}$P & A$\scriptstyle{\rightarrow}$C & A$\scriptstyle{\rightarrow}$S & 
 C$\scriptstyle{\rightarrow}$P & C$\scriptstyle{\rightarrow}$A & C$\scriptstyle{\rightarrow}$S & 
 S$\scriptstyle{\rightarrow}$P & S$\scriptstyle{\rightarrow}$A & S$\scriptstyle{\rightarrow}$C &
 Avg.\\
\midrule
DCG w/o style conf. & 67.82 & 60.95 & 56.36 & 94.83 & 67.42 & 57.70 & 75.64 & 73.46 & 63.08 & 66.66 & 60.57 & 64.06 & 67.38 \\
DCG w/o label dive. & 71.45 & 63.46 & 63.78 & \textbf{95.81} & 66.14 & 64.31 & \textbf{90.86} & \textbf{78.58} & 71.50 & 50.48 & 50.21 & 55.40 & 68.50 \\
DCG w/o dual cali. & 74.12 & 66.74 & 61.73 & 94.91 & \textbf{72.21} & 66.71 & 75.96 & 77.82 & 71.02  & 63.65 & \textbf{67.90} & 72.16 & 72.07 \\
Ours & \textbf{75.52} & \textbf{70.98} & \textbf{64.01} & 94.94  & 71.82  & \textbf{67.18} & 84.64 & 77.39 & \textbf{72.24} & \textbf{67.21} & 66.86 & \textbf{72.83} & \textbf{73.80}  \\
\bottomrule  
\end{tabular}}
\caption{Ablation studies in terms of different components in DCG module on \textbf{PACS} dataset.}
\label{tab:ablation_dcg}
\end{table*}
\begin{table*}[ht!] 
\centering
\footnotesize
\resizebox{1.0\textwidth}{!}{
\begin{tabular}{l|c|c|ccc|ccc|ccc|ccc|c}
\toprule
Method & $\mathcal{C}$ & $\mathcal{R}$ & P$\scriptstyle{\rightarrow}$A & P$\scriptstyle{\rightarrow}$C & P$\scriptstyle{\rightarrow}$S & 
 A$\scriptstyle{\rightarrow}$P & A$\scriptstyle{\rightarrow}$C & A$\scriptstyle{\rightarrow}$S & 
 C$\scriptstyle{\rightarrow}$P & C$\scriptstyle{\rightarrow}$A & C$\scriptstyle{\rightarrow}$S & 
 S$\scriptstyle{\rightarrow}$P & S$\scriptstyle{\rightarrow}$A & S$\scriptstyle{\rightarrow}$C &
 Avg.\\
\midrule
\multirow{4}{*}{MixUp} & 1 & 0.4  & 73.60 & 67.98 & 61.35 & 94.71 & 73.11 & 66.42 & 85.90 & 75.22 & 68.72 & 61.67 & 61.79 & 67.21 & 71.47\\
& 3 & 0.4 & 75.52 & \textbf{70.98} & \textbf{64.01} & 94.94 & 71.82 & 67.18 & 84.64 & \textbf{77.39} & \textbf{72.24} & \textbf{67.21} & \textbf{66.86} & \textbf{72.83} & \textbf{73.80} \\
& 5 & 0.4 & \textbf{75.68} & 67.43 & 63.96 & \textbf{95.51} & \textbf{73.45} & 67.62 & 87.69 & 77.08 & 71.29 & 65.25 & 63.73 & 68.97 & 73.14 \\
& 10 & 0.4 & 75.16 & 66.81 & 63.99 & 95.48 & 72.79 & \textbf{67.71} & \textbf{88.54} & 77.03 & 71.82 & 64.72 & 61.43 & 67.18 & 72.72 \\
\midrule
\multirow{4}{*}{MixUp} & 3 & 0.2 & 72.46 & 69.87 & 63.98 & 94.78 & 70.88 & 67.81 & 83.22 & 75.69 & \textbf{72.79} & 66.01 & 65.83 & 71.97 & 72.94  \\
& 3 & 0.4 & 75.52 & 70.98 & \textbf{64.01} & 94.94 & \textbf{71.82} & 67.18 & 84.64 & 77.39  & 72.24 & \textbf{67.21} & 66.86 & 72.83 & \textbf{73.80} \\
& 3 & 0.6 & 74.88  & \textbf{71.08} & 62.55 & 95.13 & 69.09 & 68.31 & 82.97 & 76.64 & 72.04 & 66.10 & \textbf{68.41}& \textbf{73.52} & 73.39 \\
& 3 & 0.8 & \textbf{76.40} & 67.60 & 60.61 & \textbf{95.45} & 70.67 & \textbf{68.62} & \textbf{85.71} & \textbf{77.50} & 71.60 & 66.90 & 67.95 & 71.56 & 73.38 \\
\midrule
CutMix & 3 & 0.4 & \textbf{77.15} & \textbf{72.01} & \textbf{65.05} & 94.93  & 71.41 & 63.76 & 81.48 & 76.28 & 71.64 & 65.08 & 67.72 & \textbf{74.48} & 73.42 \\
MixUp & 3 & 0.4 & 75.52 & 70.98 & 64.01 & 94.94 & 71.82 & \textbf{67.18} & 84.64 & \textbf{77.39} & 72.24 & \textbf{67.21} & 66.86 & 72.83 & 73.80 \\
``X+U'' & 3 & 0.4 & 74.63 & 69.71 & 63.34 & \textbf{95.90} & \textbf{74.56} & 67.11 & \textbf{88.00} & 77.13 & \textbf{72.35} & 64.24 & \textbf{68.37} & 72.76 & \textbf{74.01} \\
\bottomrule  
\end{tabular}}
\vspace{-0.1cm}
\caption{Ablation studies in terms of EID on \textbf{PACS} dataset. $\mathcal{C}$ means the EID cycles, and $\mathcal{R}$ means the clean rate of EID.}
\vspace{-0.3cm}
\label{tab:ablation_eid}
\end{table*}
In this section, we first present experimental settings in detail, and then conduct plenty of ablation studies to explore the properties and effectiveness of Adversarial Pseudo Labeling (APL) module, Dual Calibrative Generalization (DCG) module and Evolving Intermediate Domain (EID), respectively. Next, we evaluate the effectiveness of our method by comparing with strong baselines like domain adaptation (DA), domain generalization (DG) and semi-supervised learning (SSL) methods on different close-set SSDG benchmarks. Further, we also evaluate our method on the newly-constructed open-set SSDG benchmarks, namely PACS-Webdata and Office-Webdata, using different settings of web-labels. To sum up, our method is evaluated both on the close-set and open-set SSDG benchmarks, and the experimental results prove the superiority of our method to other competitive methods.

\subsection{Datasets and Experimental Setup}\label{sec:exp-setup}
\subsubsection{Datasets}
Our method is evaluated on three DG datasets:
\begin{enumerate}
    \item[-] \textbf{PACS}~\cite{pacs} contains 7 categories from 4 domains: photo (\textbf{P}), art painting (\textbf{A}), cartoon (\textbf{C}), and sketch (\textbf{S}). We follow the train-test split strategy in the prior work~\cite{pacs} to split the training domains into 9:1 (train:val) where only the train split can be used to train model. 
    \item[-] \textbf{Digits-DG}~\cite{ddaig} includes 4 domains: MNIST (\textbf{Mn}), MNIST-m (\textbf{Mm}), SVHN (\textbf{Sv}), and Synthetic digits (\textbf{Sy}). Each domain contains $600$ images per class sampled from MNIST~\cite{mnist}, MNIST-m~\cite{dann}, SVHN~\cite{svhn} and Synthetic-digits~\cite{SynDigits} datasets, respectively. Following the prior work~\cite{ddaig}, we split the training domains into 8:2 (train:val).
    \item[-] \textbf{Office-Home}~\cite{office-home} is a medium-size DG benchmark containing $15500$ images with $65$ categories from 4 distinct domains: Artistic images (\textbf{Ar}), Clip Art (\textbf{Cl}), Product images (\textbf{Pr}), and Real-world iamges (\textbf{Rw}). These domains are varied in background, viewpoint and image style. 
\end{enumerate}

\subsubsection{Experimental Setup}
For close-set SSDG, we conduct leave-one-domain-out strategy in each dataset that selects three source domains for training and leaves the remaining one for evaluation. Specifically, the three source domains consists of one labeled domain and two unlabeled domain, which can form 12 kinds of SSDG tasks per dataset. We evaluate our method on all the SSDG tasks for each dataset and report the average accuracy. Note that in each experiment we report the average of the last five epochs as the final result.

For open-set SSDG, we follow most of close-set SSDG settings, and the main difference is to use web-crawled data as the unlabeled source domain. Therefore, the source domains contain two parts: labeled source domain part and web-crawled source domain part. 
However, unlike the unlabeled data in close-set SSDG, the web-crawled part contains vague annotation, referred to as web-label. We devise multiple approaches to utilize these web-label, which will be introduced later.

\subsection{Implementation Details}
All experiments are conducted on PyTorch $1.7.1$ platform with RTX 3090 GPU. For each SSDG task, we train domain-specific APL for each unlabeled domain, to avoid the disaster of performance degradation brought by training the labeled domain and multiple unlabeled domains simultaneously. Hence, there involves two APL modules and a DCG module in each experiment. Without any specific statement, we use MCD to implement APL~\cite{mcd} and use MixUp to implement EID by default. We use ResNet-18~\cite{resnet} as the backbone for APL and DCG, which is initialized with the weights pretrained on ImageNet~\cite{deng2009imagenet}. For DCG module, the style confusor layers are inserted into the feature maps of the $1^{th}$, $2^{rd}$, and $3^{rd}$ residual blocks of each subnetwork. The APL and DCG are trained with SGD, batch size of 128, initial learning rate of 1e-3, weight decay of 5e-4, momentum of 0.9, and maximum epochs of 30 and 15, respectively. The learning rate is fixed when training each module, while it is decreased globally as the EID cycles increased, as defined by: $lr = lr_{initial}/ {T^2}$, where ${T}$ denotes the cycle index. The maximum cycles is set 3. The global clean rate in EID is set 0.4, while the local clean rate in DCG is decreased from 1 to 0.5 linearly. Note that the clean rate of EID is 0.6 for PACS-Webdata due to the noise web data. As for the input, we resize each image to $224\times 224$ and use simple augmentation techniques like random translation and random horizontal flipping on PACS and Office-Home, while do not use any augmentation techniques on Digits-DG where each image is resized to $32 \times 32$.

\subsection{Ablation Studies}
We conduct extensive ablations studies to validate the effectiveness of APL, DCG and EID, respectively. Note that these experiments are implemented with closet-set SSDG setting.
\subsubsection{Ablation Studies on APL}
\begin{itemize}[leftmargin=12pt, topsep=2pt, itemsep=0pt]
\item{\textbf{Explore Different Pseudo Labeling Methods.}}
We study the impact of APL on SSDG by using different pseudo labeling (PL) techniques to implement APL, including DANN~\cite{dann}, MCD~\cite{mcd}, and MME~\cite{MME}. As shown in Fig.\ref{fig:da-compare}(a), the accuracy of the pseudo labels is increased as the training goes on. This is because the domain gap in APL is getting smaller as EID evolves. Both the experimental results of MCD and MME is better than that of DANN, because MCD and MME can obtain higher-quality pseudo labels than DANN.
Connecting Fig.\ref{fig:da-compare}(b) to Table \ref{tab:ablation_apl}, we can see that the accuracy of pseudo labels in APL is positively related with DG performance in DCG.
\begin{figure}[t]
     \centering
     \footnotesize
     \begin{subfigure}[b]{0.24\textwidth}
         \centering
         \includegraphics[width=\textwidth]{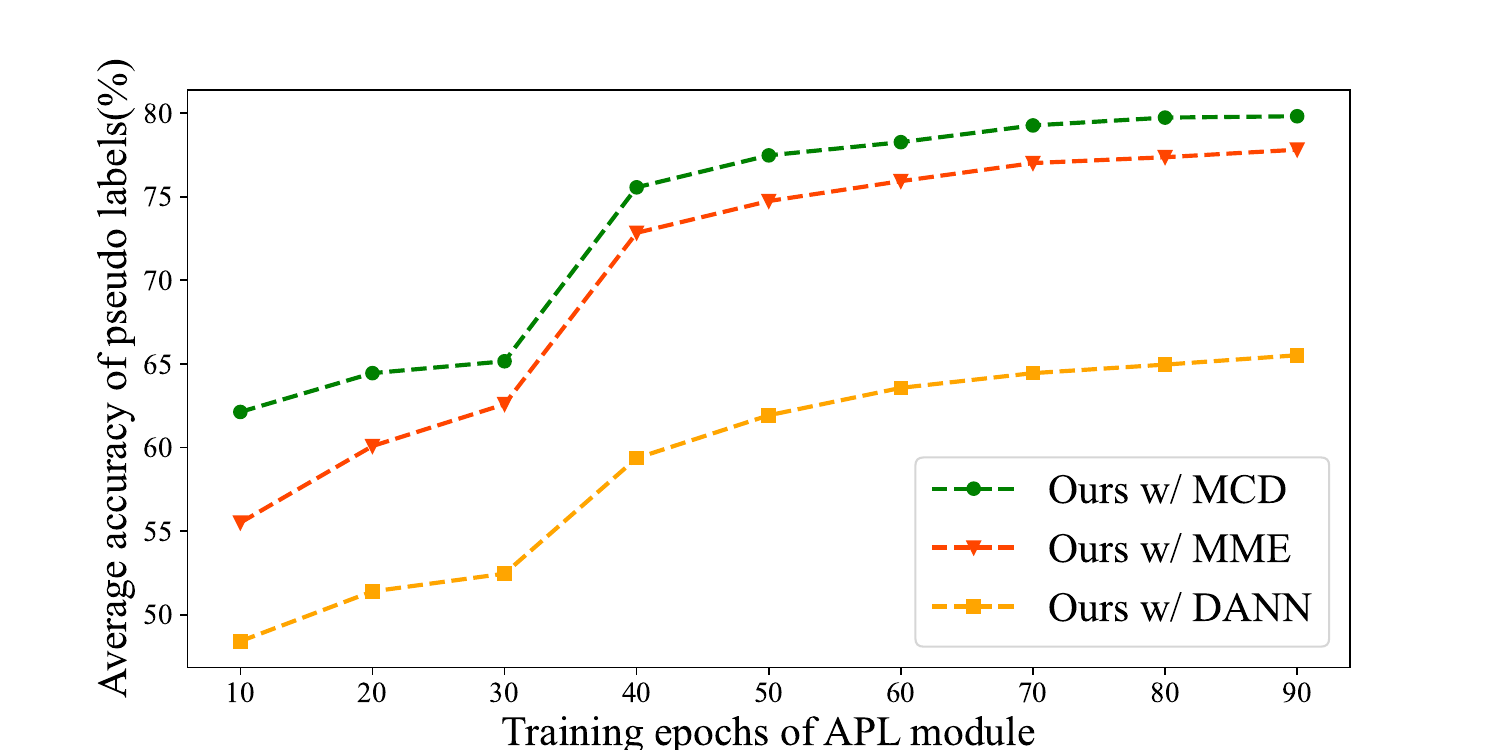}
         \caption{}
     \end{subfigure}
     \hfill
     \begin{subfigure}[b]{0.24\textwidth}
         \centering
         \includegraphics[width=\textwidth]{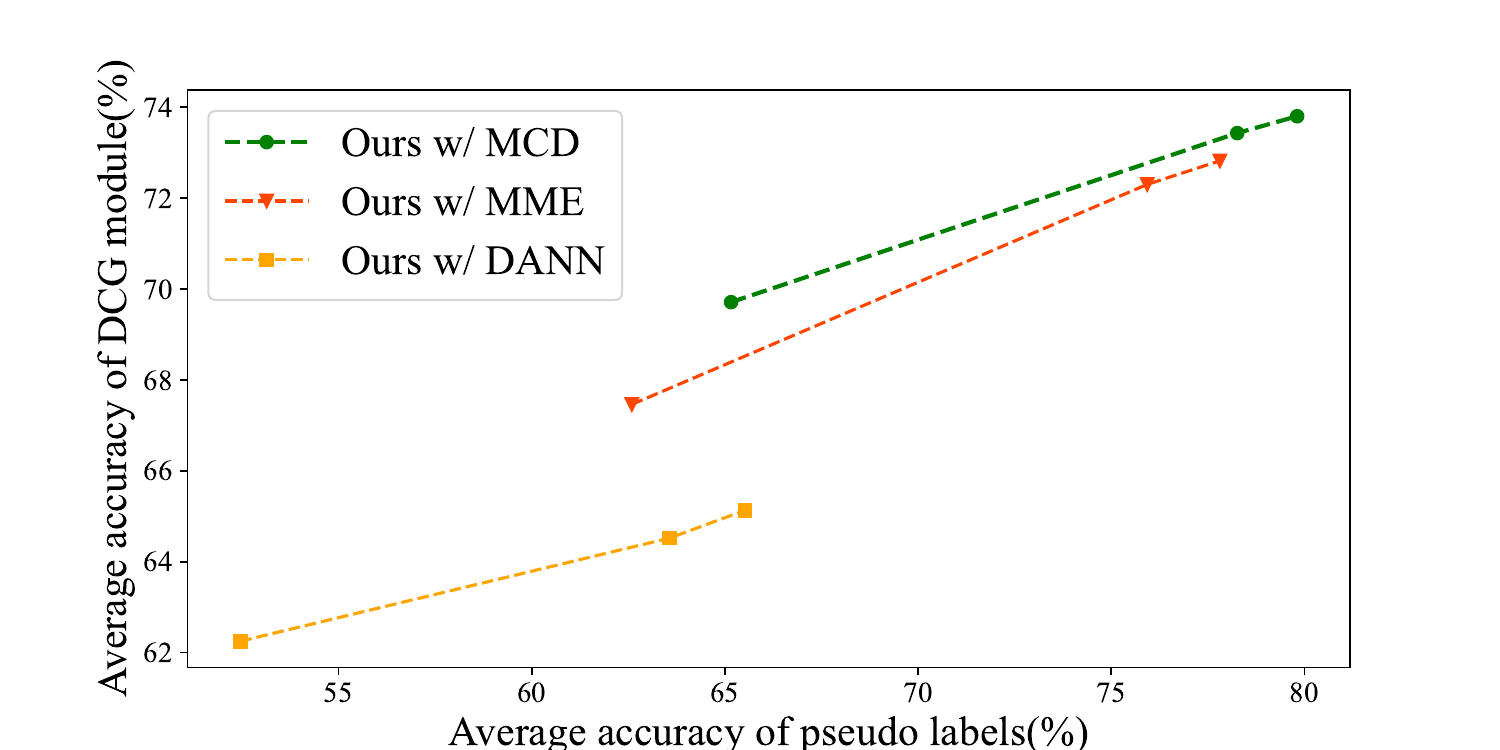}
         \caption{}
     \end{subfigure}
        \caption{(a) Average accuracy of pseudo labels produced by APL module implemented with different DA methods during training. (b) Relationship between pseudo labeling ability of APL module and generalization ability of DCG module.}
        \label{fig:da-compare}
\end{figure}
\item{\textbf{Explore Different Backbones of APL.}} 
We also explore the influence of different backbones on the APL module, by implementing the APL module with different backbones (ResNet-18, ResNet-50 and ResNet-101) while leaving the DCG module backbone unchanged (ResNet-18). When using ResNet-101 as the backbone of the APL module, the batch size is reduced to $64$ due to the limitation of GPU memory. The experimental results are shown in Table \ref{tab:ablation_apl}. From this, it can be seen that the deeper the network of the APL module, the better the generalization performance of the DCG module.
The APL module with the deepest backbone (ResNet-101) makes the DCG module achieve the best generalization performance. The reason behind this is that the APL functions as an teacher model for DCG, and a deeper network can impart more valuable knowledge to teach DCG. Furthermore, during the inference stage, the APL is discarded and only the DCG is retained, meaning that deepening APL will improve the performance of the DCG without incurring any additional inference costs. Hence, it is suggested that the APL module can be deepened to improve DCG as long as training resources are available.
\end{itemize}

\begin{figure}[!t]
     \centering
     \begin{subfigure}[b]{0.3\linewidth}
         \centering
         \includegraphics[width=\linewidth]{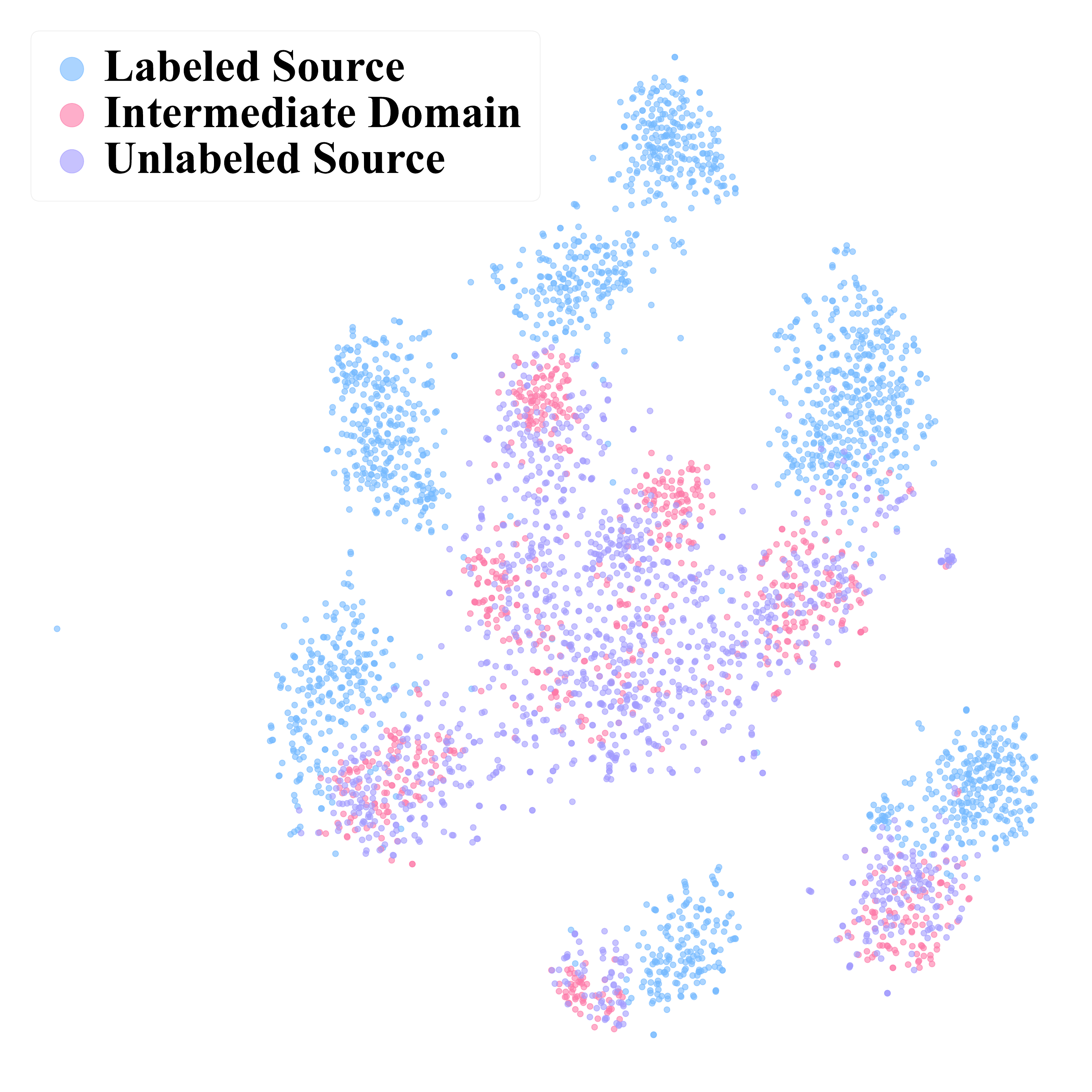}
         \caption{APL, Cycle 1}
     \end{subfigure}
     \hfill
     \begin{subfigure}[b]{0.3\linewidth}
         \centering
         \includegraphics[width=\linewidth]{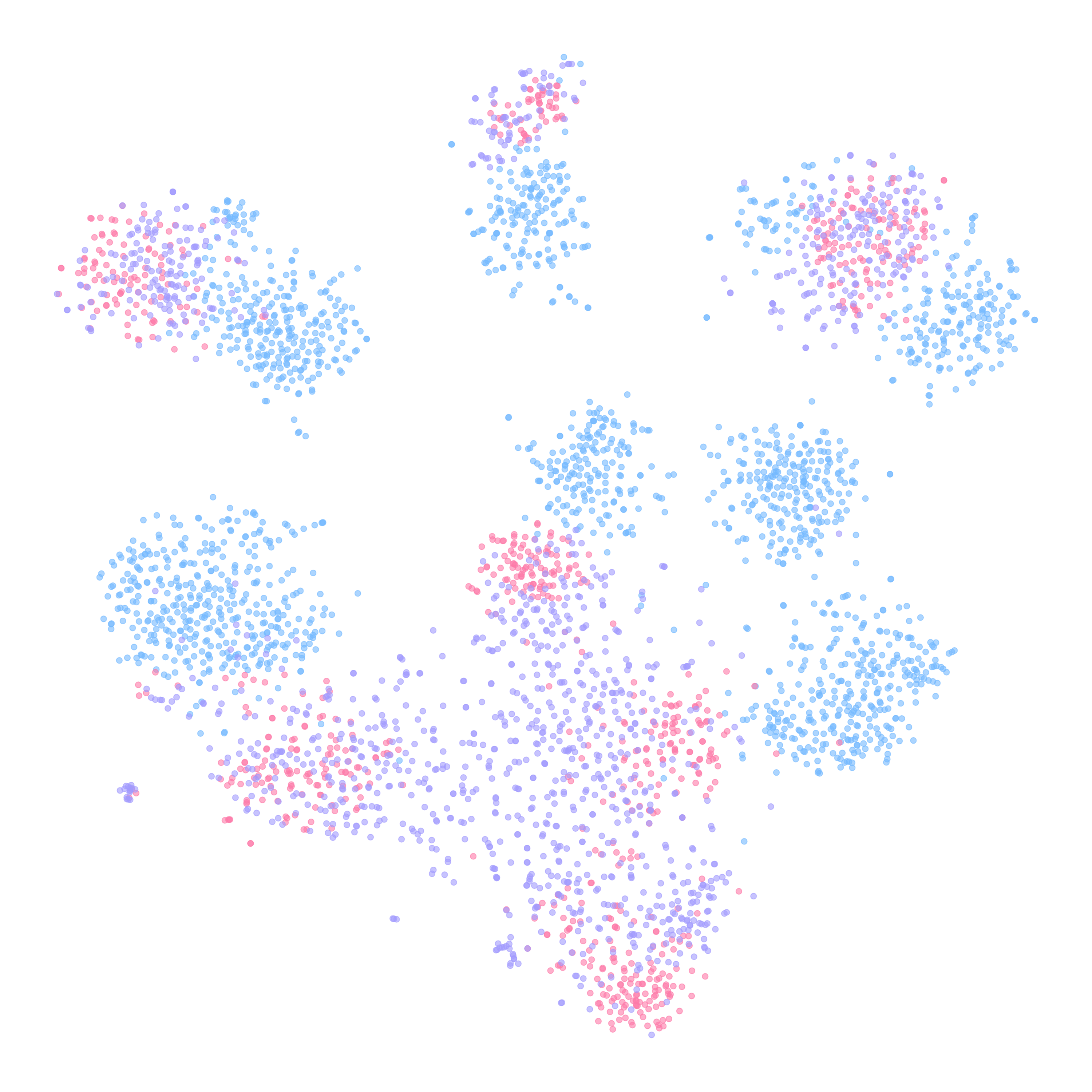}
         \caption{APL, Cycle 2}
     \end{subfigure}
     \hfill
     \begin{subfigure}[b]{0.3\linewidth}
         \centering
         \includegraphics[width=\linewidth]{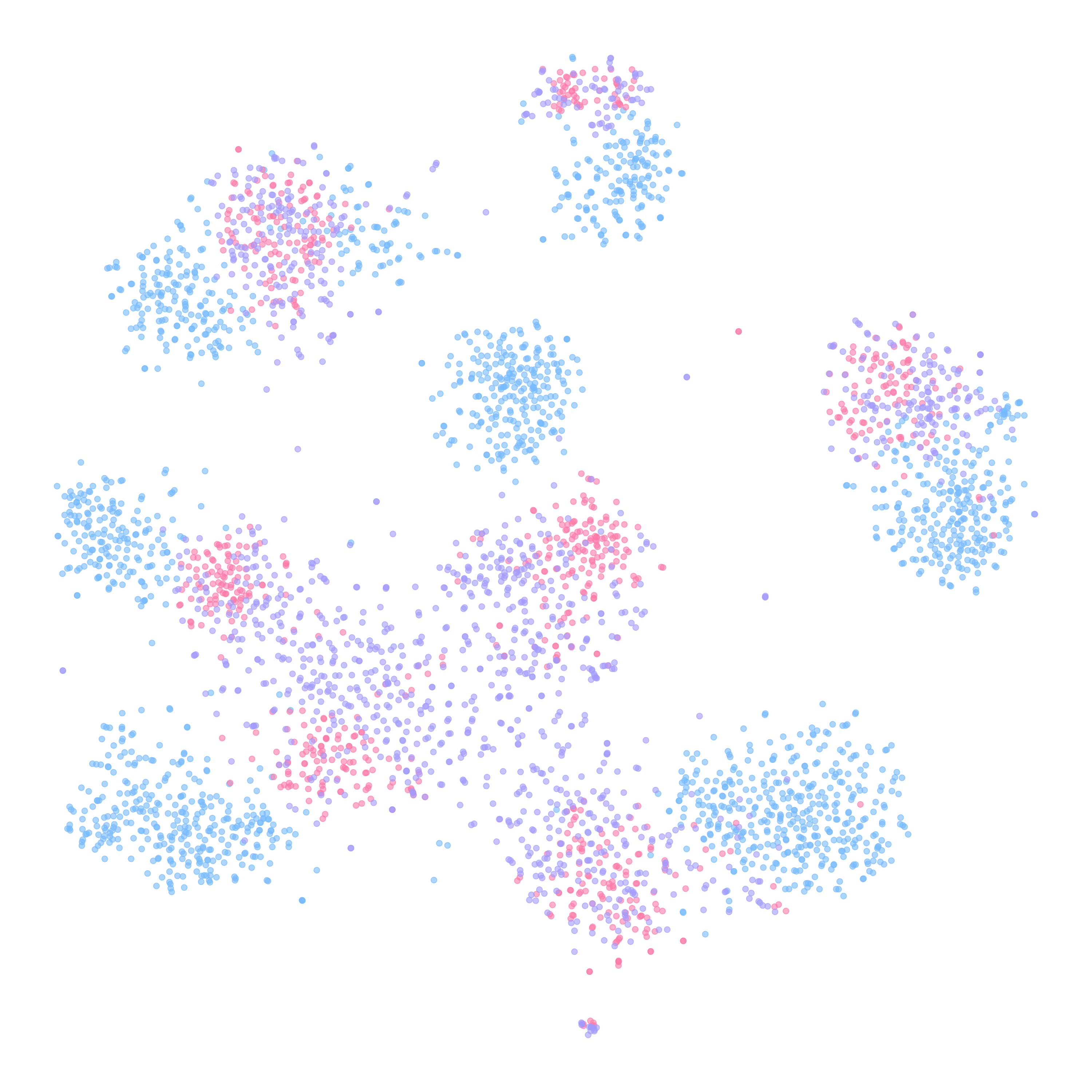}
         \caption{APL, Cycle 3}
     \end{subfigure}
     \begin{subfigure}[b]{0.3\linewidth}
         \centering
        \includegraphics[width=\linewidth]{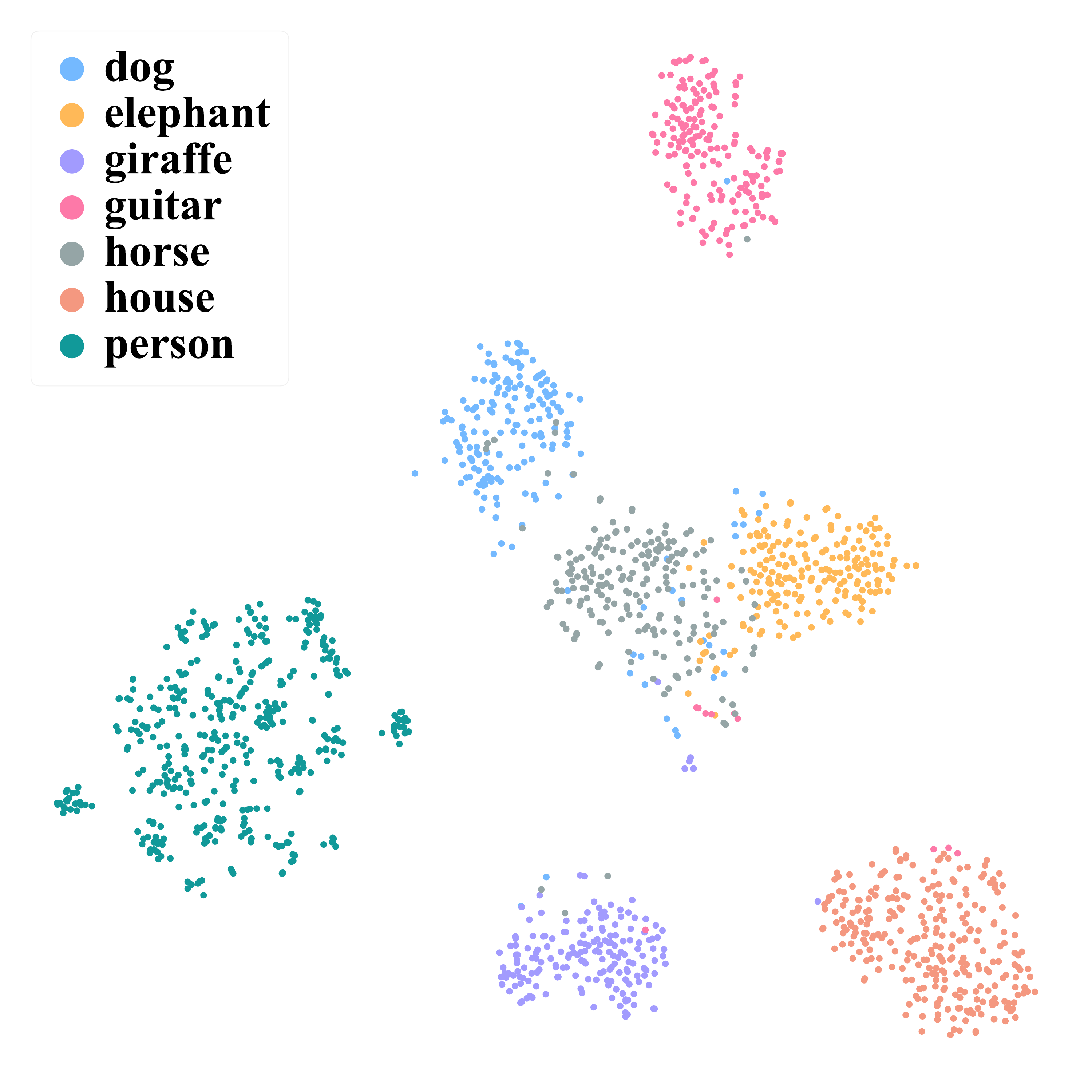}
         \caption{DCG, Cycle 1}
     \end{subfigure}
     \hfill
     \begin{subfigure}[b]{0.3\linewidth}
         \centering
        \includegraphics[width=\linewidth]{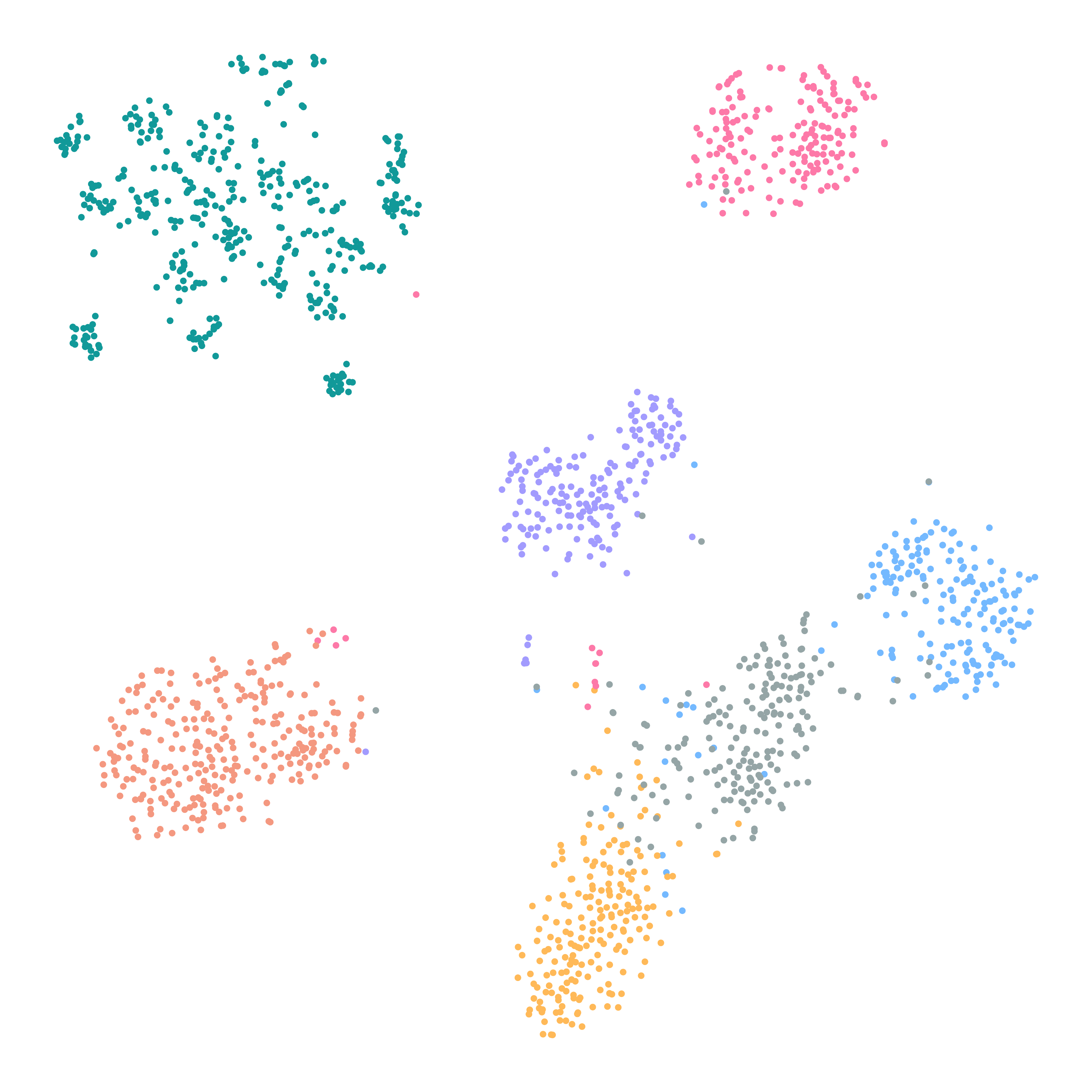}
         \caption{DCG, Cycle 2}
     \end{subfigure}
     \hfill
     \begin{subfigure}[b]{0.3\linewidth}
         \centering
        \includegraphics[width=\linewidth]{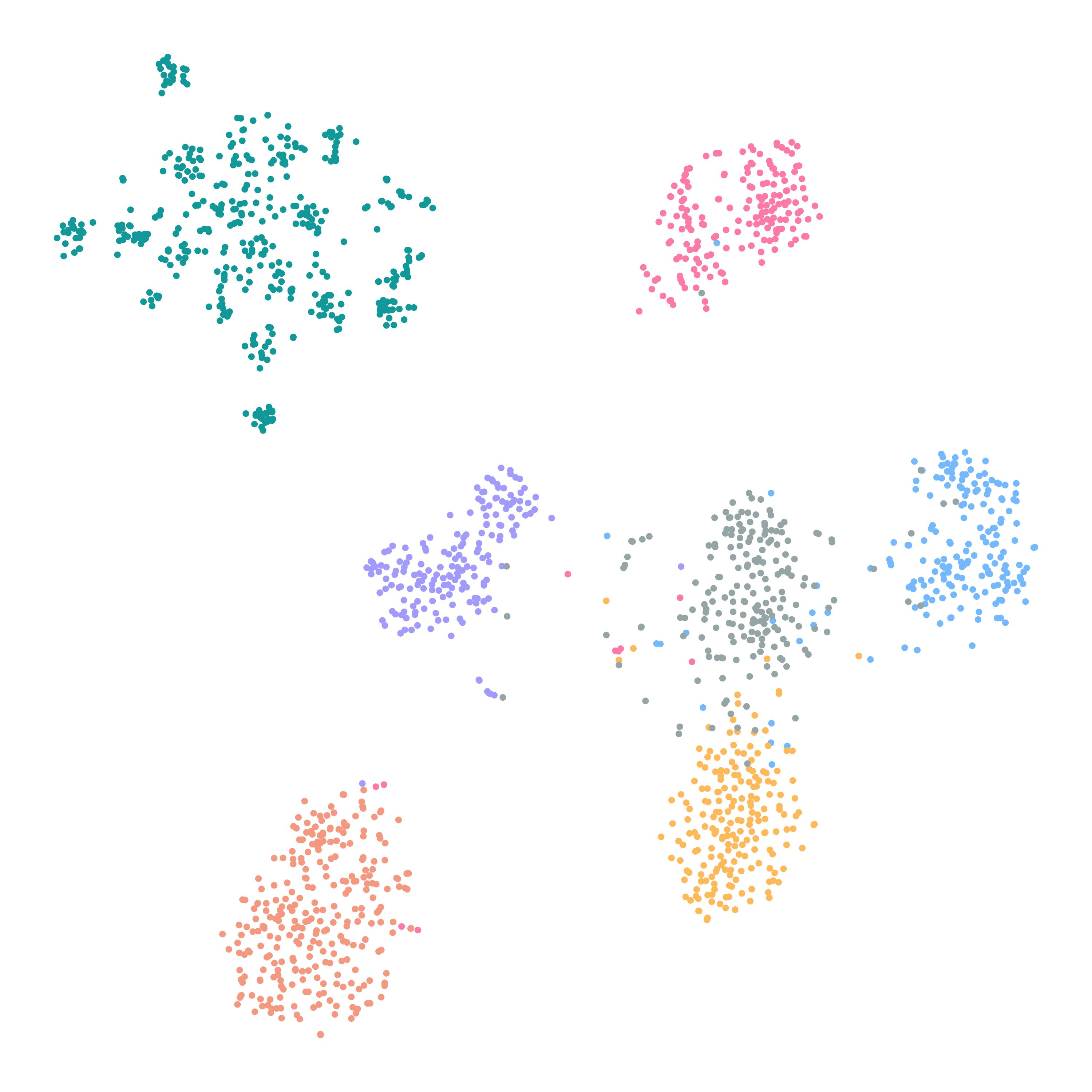}
         \caption{DCG, Cycle 3}
     \end{subfigure}
        \caption{t-SNE visualization of feature distributions predicted by APL and DCG on PACS dataset. \textbf{Top}: The feature distributions of labeled source (Art), unlabeled source (Cartoon), and intermediate domain predicted by APL in different cycles. \textbf{Bottom}: The feature distributions of unseen target domain (Photo) predicted by DCG in different cycles. }
        \label{fig:visualize}
\end{figure}
\begin{table*}[ht]  
\centering
\footnotesize
\resizebox{1.0\textwidth}{!}{
\begin{tabular}{l|l|ccc|ccc|ccc|ccc|c}  
\toprule
Type & Method & P$\scriptstyle{\rightarrow}$A & P$\scriptstyle{\rightarrow}$C & P$\scriptstyle{\rightarrow}$S & 
 A$\scriptstyle{\rightarrow}$P & A$\scriptstyle{\rightarrow}$C & A$\scriptstyle{\rightarrow}$S & 
 C$\scriptstyle{\rightarrow}$P & C$\scriptstyle{\rightarrow}$A & C$\scriptstyle{\rightarrow}$S & 
 S$\scriptstyle{\rightarrow}$P & S$\scriptstyle{\rightarrow}$A & S$\scriptstyle{\rightarrow}$C &
 Avg.\\ 
\midrule
\multirow{4}{*}{DA} & DANN~\cite{dann}  & 58.89 & 42.70 & 19.83 & 93.35 & 69.07 & 48.27 & 86.47 & 64.65 & 64.82 & 40.30 & 36.52 & 51.11 & 56.33 \\
 & CDAN+E~\cite{cdan} & 59.72 & 60.11 & 27.65 & 90.48 & \textbf{75.51} & 51.81 & 83.53 & 57.03 & 62.02 & 35.69 & 52.59 & 57.12 & 59.44 \\
 & MCD~\cite{mcd}  & 39.45 & 52.01 & 19.42 & 80.06 & 74.02 & 62.65 & 59.46 & 52.59 & 60.97 & 17.43  & 52.15 & 49.79 & 51.67 \\
 & MME~\cite{MME} & 59.42 & 50.60 & 23.80 & 92.10 & 72.27 & 56.62 & 83.59 & 54.15 & 65.15 & 36.11 & 29.49 & 53.71 & 56.42 \\ 
 \midrule
\multirow{4}{*}{DG} & RSC~\cite{huang2020rsc} & 66.59 & 27.62 & 38.55 & 93.71 & 68.03 & 65.69 & 83.51 & 69.17 & \textbf{76.62} & 47.50 & 43.00 & 65.19 & 62.10 \cr
 & L2D~\cite{l2d} & 65.17 & 30.72 & 35.39 & 96.10  & 65.68 & 57.97 & 87.34 & 73.45 & 67.93 & 48.24 & 45.86 & 61.77 & 61.30 \cr
 & DGvGS~\cite{DGvGs} & 54.20 & 16.60 & 28.51 & 93.83 & 54.74 & 39.71 & 80.30 & 59.47 & 56.73 & 14.31 & 16.21 & 17.19 & 46.78  \cr
\midrule
\multirow{3}{*}{\tabincell{c}{DA\\+\\DG}} & MCD+DDAIG\cite{ddaig} & 70.95 & 52.97 & 55.00 & 93.57 & 64.71 & 65.18 & 90.32 & 74.47 & 69.73 & 47.03 & 45.91 & 45.15 & 64.58  \cr
 & MCD+Mixstyle~\cite{mixstyle} & \textbf{75.80} & 61.19 & 50.81 & 95.09 & 67.87 & 64.03 & 89.81 & \textbf{78.44} & 68.30 & 48.38 & 44.60 & 53.23 & 66.46 \cr
 & ``CDAN+E''+DAELDG~\cite{DAELDG} & 62.32 & 62.32 & 31.9  & 95.59 & 63.53 & 46.58 & 85.97 & 73.58 & 69.99 & 38.99 & 38.23 & 55.06 & 60.34 \cr
\midrule
\multirow{4}{*}{SSL} & MeanTeacher~\cite{Mean-teacher} & 54.70 & 36.26 & 33.09 & 91.90 & 65.59 & 38.07 & 80.26 & 60.59 & 58.60 & 38.08 & 33.38 & 54.70 & 53.77  \\
 & MixMatch~\cite{MixMatch} & 35.73 & 16.20 & 24.53 & 87.25 & 62.67 & 47.57 & 43.07 & 47.85 & 50.65 & 26.10 & 46.86 & 52.17 & 45.05 \\
 & FixMatch~\cite{FixMatch} & 66.82 & 34.92 & 25.88 & \textbf{96.61} & 72.93 & 67.14 & \textbf{91.72} & 76.50 & 69.54 & 36.30 & 35.19 & 55.96 & 60.79 \\
\midrule
\multicolumn{2}{c|}{Ours w/ MixUp}  & 75.52 & \textbf{70.98} & \textbf{64.01} & 94.94  & 71.82  & \textbf{67.18} & 84.64 & 77.39 & 72.24 & \textbf{67.21} & 66.86 & \textbf{72.83} & 73.80   \cr
\multicolumn{2}{c|}{Ours w/ ``X+U''} & 74.63 & 69.71 & 63.34 & 95.90 & 74.56 & 67.11 & 88.00 & 77.13 & 72.35 & 64.24 & \textbf{68.37} & 72.76 & \textbf{74.01} \cr
\bottomrule  
\end{tabular}}
\caption{The generalization performance of our method in comparison with DA, DG and SSL methods on \textbf{PACS} dataset.}
\vspace{-0.1cm}
\label{tab:pacs}
\end{table*}
\begin{table*}[ht]  
\centering
\footnotesize
\resizebox{1.0\textwidth}{!}{
\begin{tabular}{l|l|ccc|ccc|ccc|ccc|c}  
\toprule
Type & Method & Mn$\!\scriptstyle{\rightarrow}\!$Mm & Mn$\!\scriptstyle{\rightarrow}\!$Sv & Mn$\!\scriptstyle{\rightarrow}\!$Sy & 
 Mm$\!\scriptstyle{\rightarrow}\!$Mn & Mm$\!\scriptstyle{\rightarrow}\!$Sv & Mm$\!\scriptstyle{\rightarrow}\!$Sy & 
 Sv$\!\scriptstyle{\rightarrow}\!$Mn & Sv$\!\scriptstyle{\rightarrow}\!$Mm & Sv$\!\scriptstyle{\rightarrow}\!$Sy & 
 Sy$\!\scriptstyle{\rightarrow}\!$Mn & Sy$\!\scriptstyle{\rightarrow}\!$Mm & Sy$\!\scriptstyle{\rightarrow}\!$Sv &
 Avg.\\
\midrule
\multirow{4}{*}{DA} & DANN~\cite{dann}  & 41.52 & 19.32 & 44.83 & 97.38 & 37.98 & 52.12 & 67.48 & 43.73 & 87.58 & 82.93 & 45.47 & 68.85 & 57.43 \cr
& CDAN+E~\cite{cdan}  & 44.65 & 14.42 & 49.78 & 96.85 & 38.00    & 55.60  & 62.90  & 51.62 & 86.12 & 85.57 & 56.13 & 69.20  & 59.24 \cr
& MCD~\cite{mcd}  & 48.45 & 23.85 & 42.73 & 97.82 & 36.38 & 58.00    & 64.30  & 49.48 & 85.65 & 88.28 & 57.50 & 72.13 & 60.38 \cr
& MME~\cite{MME}   & 40.75 & 20.02 & 47.47 & \textbf{97.87} & 40.50  & 60.53 & 69.43 & 50.62 & 87.07 & 84.62 & 49.88 & 69.97 & 59.89  \cr
\midrule
\multirow{3}{*}{DG} & RSC~\cite{huang2020rsc} & 42.80  & 19.32 & 45.00    & 93.63 & 11.68 & 12.03 & 70.47 & 46.13 & \textbf{95.45} & 81.36 & 42.36 & 78.82 & 53.25  \cr
& L2D~\cite{l2d}& \textbf{57.22} & 28.15 & 53.10 & 97.09 & 12.48 & 25.07 & 72.43 & 52.78 & 94.24 & 80.24 & 45.66 & \textbf{79.95} & 58.20 \cr
& DGvGS~\cite{DGvGs}  & 13.33 & 12.17 & 19.45 & 88.88 & 11.20 & 16.47 & 57.68 & 24.20 & 88.50 & 68.37 & 25.87 & 67.07 & 41.10  \cr
\midrule
\multirow{3}{*}{\tabincell{c}{DA\\+\\DG}} & MCD+DDAIG~\cite{ddaig}  & 34.22 & 16.83 & 33.75 & 95.05 & 29.93 & 53.59 & 64.46 & 39.70  & 83.52 & 73.44 & 46.10  & 59.07 & 52.47  \cr
 & MCD+Mixstyle~\cite{mixstyle} & 45.36 & 24.74 & 48.34 & 96.93 & 36.71 & 56.61 & 66.49 & 42.86 & 84.13 & 75.02 & 49.29 & 67.29 & 57.81 \cr
 & ``CDAN+E''+DAELDG~\cite{DAELDG} & 41.21 & 17.24 & 40.63 & 93.80  & 34.57 & 52.39 & 52.72 & 42.28 & 84.19 & 74.59 & 47.89 & 46.65 & 52.35  \cr
 \midrule
\multirow{4}{*}{SSL}
& MeanTeacher~\cite{Mean-teacher}& 23.91 & 13.80  & 26.43 & 82.13 & 19.30  & 32.42 & 43.59  & 17.49 & 59.19 & 56.28 & 22.16 & 38.36 & 36.26 \cr
& MixMatch~\cite{MixMatch} & 32.99 & 18.09 & 30.32 & 93.61 & 26.69 & 45.42 & 59.25  & 27.93 & 76.68 & 67.71 & 36.89 & 51.54 & 47.26 \cr
& FixMatch~\cite{FixMatch} & 29.89 & 10.63 & 23.88 & 90.84 & 32.46 & 48.18 & 57.51  & 40.03 & 70.93 & 73.95 & 51.94 & 61.34 & 49.30 \cr
\midrule
\multicolumn{2}{c|}{Ours w/ MixUp} & 51.56 & \textbf{37.29} & \textbf{53.30} & 97.12 & \textbf{58.60} & 69.05 & \textbf{87.73}  & \textbf{60.89} & 87.51 & \textbf{92.39} & \textbf{64.21} & 70.89 & \textbf{69.21} \cr
\multicolumn{2}{c|}{Ours w/ ``X+U''} & 52.00 & 30.91  & 50.53 & 97.45 & 56.57  & \textbf{71.08} & 81.95  & 58.90 & 86.52 & 88.08 & 62.27 & 71.49 & 67.31 \cr
\bottomrule  
\end{tabular}  }
\vspace{-0.1cm}
\caption{The generalization performance of our method in comparison with DA, DG and SSL method on \textbf{Digits-DG} dataset.}
\vspace{-0.1cm}
\label{tab:digits}
\end{table*}
\begin{table*}[ht!]  
\centering
\footnotesize
\resizebox{1.0\textwidth}{!}{
\begin{tabular}{l|l|ccc|ccc|ccc|ccc|c}  
\toprule
Type & Method & Ar$\!\scriptstyle{\rightarrow}\!$Cl & Ar$\!\scriptstyle{\rightarrow}\!$Pr & Ar$\!\scriptstyle{\rightarrow}\!$Rw & 
Cl$\!\scriptstyle{\rightarrow}\!$Ar & Cl$\!\scriptstyle{\rightarrow}\!$Pr & Cl$\!\scriptstyle{\rightarrow}\!$Rw & 
Pr$\!\scriptstyle{\rightarrow}\!$Ar & Pr$\!\scriptstyle{\rightarrow}\!$Cl & Pr$\!\scriptstyle{\rightarrow}\!$Rw & 
Rw$\!\scriptstyle{\rightarrow}\!$Ar & Rw$\!\scriptstyle{\rightarrow}\!$Cl & Rw$\!\scriptstyle{\rightarrow}\!$Pr &
Avg.\\
\midrule
\multirow{4}{*}{DA} & DANN~\cite{dann} & 40.60  & 54.76 & 63.71 & 42.18 & 54.79 & 58.02 & 41.82 & 38.14 & 62.91 & 55.95 & 44.77 & 73.46 & 52.59 \cr
& CDAN+E~\cite{cdan}  & 42.59 & 50.03 & 61.60  & 43.47 & 54.85 & 57.79 & 41.08 & 41.42 & 61.62 & 54.47 & 45.29 & 72.40  & 52.22 \cr
& MCD~\cite{mcd}  & 42.06 & 53.95 & 62.08 & 42.89 & 55.19 & 57.59 & 39.27 & 38.67 & 64.86 & 52.66 & 43.55 & 72.58 & 52.11 \cr
& MME~\cite{MME}  & 39.40  & 52.08 & 62.45 & 46.44 & 57.42 & 58.80  & 43.76 & 41.42 & 64.86 & \textbf{56.00}    & 43.80  & 72.27 & 53.23 \cr
\midrule
\multirow{3}{*}{DG} & RSC~\cite{huang2020rsc} & 39.08 & 49.80  & 61.13 & 36.93 & 53.01 & 53.70  & 35.86 & 38.79 & 61.19 & 53.29 & 45.60  & 72.15 & 50.04  \cr
& L2D~\cite{l2d} & 39.60  & 44.84 & 57.52 & 42.20  & 52.60  & 55.65 & 38.47 & 42.95 & 62.34 & 55.03 & 48.26 & 69.33 & 50.73  \cr
& DGvGS~\cite{DGvGs} & 33.43 & 42.89 & 55.38 & 32.59 & 44.99 & 47.03 & 29.79 & 33.15 & 54.97 & 50.80 & 37.92 & 67.97 & 44.24 \cr
\midrule
\multirow{3}{*}{\tabincell{c}{DA\\+\\DG}} & MCD+DDAIG~\cite{ddaig}  & 42.49 & 54.26 & 63.57 &42.42       & 53.83 & 56.04 & 40.12 & 37.63 & 59.67 & 48.32      & 43.24 & 69.23 &50.90  \cr
 & MCD+Mixstyle~\cite{mixstyle} & 44.93 & 55.18 & 65.29 & 45.79 & 56.77 & 58.95 & 42.93 & 42.01 & 63.28 & 52.18 & 46.04 & 69.75 & 53.59 \cr
 & ``CDAN+E''+DAELDG~\cite{DAELDG} & 40.56 & 52.01 & 58.92 & 46.31 & 55.31 & 56.06 & 46.40  & 39.77 & 61.40  & 54.88 & 47.61 & 66.72 & 52.16 \cr
\midrule
\multirow{3}{*}{SSL} & MeanTeacher~\cite{Mean-teacher} & 35.06  & 50.52  & 60.76  & 39.06  & 51.38  & 53.98  & 35.77  & 34.46  & 62.02  & 54.35  & 43.36  & 72.22  & 49.41 \cr
& MixMatch~\cite{MixMatch} &40.00  & 51.80  & 62.41  & 43.21  & 57.60  & 58.91  & 42.03  & 38.49  & 63.61  & 55.54  & 43.69  & 72.43  & 52.48 \cr
& FixMatch~\cite{FixMatch} &41.39  & 55.30  & 64.44  & 44.39  & 57.80  & 57.47  & 43.96  & 42.20  & 65.83  & 57.23  & 45.01  & \textbf{73.65}  & 54.06 \cr
\midrule
\multicolumn{2}{c|}{Ours w/ MixUp} & 48.32 & 59.11  & \textbf{66.54} & 47.53 & 60.38  & 61.29 & 46.11  & 47.27 & \textbf{66.04} & 53.28 & \textbf{48.80} & 69.00 & 56.13 \cr
\multicolumn{2}{c|}{Ours w/ ``X+U''} & \textbf{48.91} & \textbf{59.52}  & 66.51 & \textbf{48.56} & \textbf{60.86}  & \textbf{63.98} & \textbf{48.12}  & \textbf{47.73} & 65.51 & 53.07 & 48.78 & 68.98 & \textbf{56.71} \cr
\bottomrule  
\end{tabular}  }
\vspace{-0.1cm}
\caption{The generalization performance of our method compared with DA, DG and SSL methods on \textbf{Office-Home} dataset.}
\vspace{-0.3cm}
\label{tab:office}
\end{table*}
\begin{table}[t]  
\centering
\footnotesize
\resizebox{1.0\linewidth}{!}{
\begin{tabular}{l|c|c}  
\toprule  
Method & Labeled source domain & Unlabeled source domain \cr
\midrule
DA & \Checkmark & \Checkmark  \cr
DG & \Checkmark & \XSolidBrush \cr
DA+DG & \Checkmark & \Checkmark   \cr
SSL & \Checkmark & \Checkmark  \cr
Ours & \Checkmark & \Checkmark  \cr
\bottomrule  
\end{tabular}}
\vspace{-0.1cm}
\caption{The training requirements of different types of methods for the close-set SSDG task.}
\vspace{-0.5cm}
\label{tab:dadgssl}
\end{table}
\begin{table*}[ht]
\centering
\footnotesize
\resizebox{1.0\textwidth}{!}{
\begin{tabular}{l|l|ccc|ccc|ccc|ccc|c}
\toprule
Web-label Strategy & Method & P$\scriptstyle{\rightarrow}$A & P$\scriptstyle{\rightarrow}$C & P$\scriptstyle{\rightarrow}$S & 
 A$\scriptstyle{\rightarrow}$P & A$\scriptstyle{\rightarrow}$C & A$\scriptstyle{\rightarrow}$S & 
 C$\scriptstyle{\rightarrow}$P & C$\scriptstyle{\rightarrow}$A & C$\scriptstyle{\rightarrow}$S & 
 S$\scriptstyle{\rightarrow}$P & S$\scriptstyle{\rightarrow}$A & S$\scriptstyle{\rightarrow}$C &
 Avg.\\ 
\midrule
\multirow{5}{*}{Web-label ignored} & MixMatch~\cite{MixMatch} &51.75 &34.69 &5.00 &94.26 &55.38 &30.65 &84.84 &68.91 &37.21 &13.89 &20.17 &31.86 &40.05  \\
 & FixMatch~\cite{FixMatch} &75.26 &32.59 &31.87 &96.36 &64.81 &55.98 &92.97 &\textbf{79.31} &67.79 &21.48 &27.34 &42.80 &57.38  \\
 & MeanTeacher~\cite{Mean-teacher}&63.45 &34.58 &33.72 &95.51 &59.12 &42.73 &85.20 &63.59 &66.18 &40.54 &41.31 &53.54 &56.62  \\
 & Ours w/ MCD & 74.55  & 46.25 & 28.71 & 96.15  & 68.98 & 40.73 & 91.12  & 73.47 & 54.47 & 65.95  & 60.93 & 62.62& 63.66 \\
 & Ours w/ OSDA & 71.07  & 42.63 & 33.81 & 96.14  & 63.14 & 43.05 & 84.74  & 71.72 & 59.41 & 57.47  & 52.18 & 58.45& 61.15 \\
\midrule
\multirow{5}{*}{Web-label retained}& RSC~\cite{huang2020rsc} & 60.00 & 63.75 & 53.87 & 82.65 & 68.01 & 59.39 & 81.85 & 65.92 & 73.87 & 72.74 & 59.12 & 66.62 & 67.32 \\
& MixStyle~\cite{mixstyle}& 67.38 & 58.14 & 44.73 & 90.08 & 67.52 & 55.79 & 92.89 & 71.78 & 62.70  & 90.27 & 66.65 & 67.91 & 69.65 \\
& L2D~\cite{l2d}& 66.86 & 64.60  & 57.20  & 83.81 & 69.44 & 61.24 & 79.52 & 67.64 & 66.43 & 76.23 & 63.35 & 70.58 & 68.91 \\
& DGvGS~\cite{DGvGs}& 60.45 & 43.13 & 29.60 & 95.27 & 71.50 & 50.32 & 91.68 & 71.68 & 63.94 & 79.04 & 50.39 & 49.53 & 64.27 \\
& Ours w/o APL  & 72.21 & 68.30 & 55.08 & 97.02  & 74.73  & \textbf{62.26} & 95.13  & 74.12 & 68.11 & \textbf{95.63}  & 70.18 & 72.50 & 75.44 \\
\midrule
\multirow{2}{*}{Web-label refined}  & Ours w/ MCD & \textbf{76.69} & \textbf{69.02} & 52.33 & \textbf{97.05}  & \textbf{75.90} & 56.55 & \textbf{95.14}  & \textbf{75.08} & \textbf{68.61}  & 95.17  & \textbf{72.81} & \textbf{73.71} & \textbf{75.67} \\
 & Ours w/ OSDA & 72.81  & 66.64 & \textbf{58.20} & 96.75  & 75.91 & 49.09 & 93.43  & 72.84 & 65.23 & 92.05  & 70.35 & 75.01& 74.03 \\
\bottomrule
\end{tabular}}
\vspace{-0.1cm}
\caption{The generalization performance of our method in comparison with other methods on \textbf{PACS-Webdata} dataset.}
\vspace{-0.1cm}
\label{tab:pacs_web}
\end{table*}
\begin{table*}[ht]
  \centering
\resizebox{1.0\textwidth}{!}{
    \begin{tabular}{l|l|ccc|ccc|ccc|ccc|c}
    \toprule
 Web-label Strategy & Method & Pr$\!\scriptstyle{\rightarrow}\!$Ar  & Pr$\!\scriptstyle{\rightarrow}\!$Cl  & Pr$\!\scriptstyle{\rightarrow}\!$Rw  & Ar$\!\scriptstyle{\rightarrow}\!$Pr  & Ar$\!\scriptstyle{\rightarrow}\!$Cl  & Ar$\!\scriptstyle{\rightarrow}\!$Rw  & Cl$\!\scriptstyle{\rightarrow}\!$Pr  & Cl$\!\scriptstyle{\rightarrow}\!$Ar  & Cl$\!\scriptstyle{\rightarrow}\!$Rw & Rw$\!\scriptstyle{\rightarrow}\!$Pr  & Rw$\!\scriptstyle{\rightarrow}\!$Ar  & Rw$\!\scriptstyle{\rightarrow}\!$Cl & Avg. \\
    \midrule
 \multirow{2}{*}{Web-label ignored}   & Ours w/ MCD  & 47.08        & 46.19        & 70.00           & 62.90         & 47.41        & 68.06        & 65.38        & 49.37        & 66.34        & 75.13        & 54.87        & 50.90         & 58.64 \\
& Ours w/ OSDA & 45.43        & 57.80        & 65.30           & 46.32         & 58.82        & 61.45        & 43.08       & 44.92        & 67.92        & 54.88        & 49.64        & 73.41         & 55.75 \\
\midrule
\multirow{1}{*}{Web-label retained}  & Ours w/o APL  & 44.94        & 48.48        & 69.67        & 64.72        & 49.15        & 69.28        & 69.25        & 47.97        & 68.17        & 73.66        & 52.27        & 50.67        & 59.02 \\
\midrule
 \multirow{2}{*}{Web-label refined} & Ours w/ MCD    & 47.48 & 49.13 & \textbf{72.15} & \textbf{68.16} & 49.79 & \textbf{73.14} & \textbf{70.62} & \textbf{50.70} & 70.31 & \textbf{76.52} & \textbf{55.61} & 53.33 & \textbf{61.41} \\
    & Ours w/ OSDA   & \textbf{49.63} & \textbf{68.59} & 71.71 &48.64 & \textbf{69.96} & 69.45 & 46.94 & 48.43 & \textbf{71.60} & 56.25 & 52.66 & \textbf{75.93} & 60.82 \\
    \bottomrule
    \end{tabular}%
}
\vspace{-0.1cm}
  \caption{The generalization performance of our method on \textbf{Office-Webdata} dataset}
  \label{tab:office_web}%
\vspace{-0.3cm}
\end{table*}
\begin{table}[ht]  
\centering
\footnotesize
\resizebox{0.99\linewidth}{!}{
\begin{tabular}{l|c|c|l}  
\toprule  
\multirow{2}{*}{Web-label strategy} &\multirow{2}{*}{Labeled source domain} & \multicolumn{2}{c}{Web-crawled data} \cr
\cline{3-4}
& &  Web image & Web label \cr
\midrule
Web-label ignored & \Checkmark & \Checkmark  & \XSolidBrush \cr
Web-label retained & \Checkmark & \Checkmark & \Checkmark (Unchanged) \cr
Web-label refined & \Checkmark & \Checkmark   & \Checkmark (Refined) \cr
\bottomrule  
\end{tabular}}
\vspace{-0.1cm}
\caption{The training data of different web-label strategies.}
\vspace{-0.4cm}
\label{tab:opensetssdg}
\end{table}

\subsubsection{Ablation Studies on DCG}
We explore the effectiveness of different components in DCG, including dual-calibration, style confusion training, as well as label diversity regularization. The detailed ablation studies are shown in Table \ref{tab:ablation_dcg}, from which we can see that all the components play an important role in DCG.

\subsubsection{Ablation Studies on EID}
We explore the properties of EID by conducting ablation studies with different cycle numbers $\mathcal{C}$, different clean rate $\mathcal{R}$, and different EID manners (MixUp, ``X+U'' and CutMix). The comparison results are shown in Table \ref{tab:ablation_eid}, where we can see that:

\begin{itemize}[leftmargin=12pt, topsep=2pt, itemsep=0pt]
    \item \textbf{Explore Different EID Cycles $\mathcal{C}$.} The performance is improved gradually as EID evolves, and achieves the best result when the EID cycle number reaches 3. However, the performance is degraded when EID cycle continues enlarging, which can be imputed to the progressive leakage of noisy samples in intermediate domain. Hence, we set $\mathcal{C}=3$ by default.
    \item \textbf{Explore Different Clean Rates $\mathcal{R}$.} The clean rate of the small-loss samples has positive impact on the generalization ability when it raises from 0.2 to 0.4, since the larger rate means more diverse clean set, which leads to the better evolution of intermediate domain. However, the result is degraded when the clean rate raises to 0.6. Too large clean rate will induce the leakage of noisy samples, which negatively affects the synthesis of reliable intermediate domain. Hence, we set the clean rate as 0.4 by default.
    \item \textbf{Explore Different EID Manners.} We implement three manners to achieve intermediate domain, including CutMix, MixUp and ``X+U''. We can see that EID with ``X+U'' achieves the best result on PACS benchmark, which is merely slightly better than MixUp. In fact, the intermediate domain synthesis is not limited to these manners, which encourages a further exploration in this direction for SSDG.
    \item \textbf{t-SNE Visualization.} To further explore the mechanism of EID, we visualize the feature distributions of APL and DCG modules in Fig. \ref{fig:visualize}. The top row describes the feature distributions of APL on labeled source, unlabeled source, and intermediate domain in different cycles, where we can see that the intermediate domain bridges the labeled and unlabeled source domains progressively as the training cycle runs. The bottom row depicts the feature distributions of DCG on unseen target domain in different cycles, which shows the features are getting more discriminative to reduce outliers as EID evolves. It again demonstrates the effectiveness of EID on SSDG task.
\end{itemize}

\subsection{Comparison with Strong Baselines}
To stress the effectiveness of our method, we implement extensive strong baselines under the close-set SSDG setting:
\begin{itemize}[leftmargin=12pt, topsep=2pt, itemsep=0pt]
\item \textbf{DA}: The model is trained on the labeled and unlabeled source domains simultaneously using different DA methods, and tested on unseen target domain.
\item \textbf{DG}: The model is trained merely on the labeled source domain using single DG approaches, and then tested on unseen target domain.
\item \textbf{DA+DG}: A pseudo labeling model is trained for each unlabeled source domain using different DA methods to obtain pseudo labels. Then, the pseudo-labeled and true-labeled data are used to train a DG model using different DG methods, leaving the DG model to be evaluated.
\item \textbf{SSL}: The model is trained using different semi-supervised learning methods on the labeled and unlabeled source domains, and then evaluated on the unseen target domain.
\end{itemize}

The training requirements of different baseline methods are summarized in Table \ref{tab:dadgssl}. And our method is compared with these baseline methods on PACS, Digits-DG, and Office-Home datasets under the close-set SSDG setting. The comparison results are reported in Table \ref{tab:pacs}, Table \ref{tab:digits}, and Table \ref{tab:office}, respectively, where we can see that our method is not only superior to the existing DA, DG, SSL methods, but also outperforms the simple combination of DA and DG by a large margin, especially on PACS and Digits-DG datasets. The superiority of our method can be owned to the following parts: 1) the DCG module can facilitate to learn generalization from noisy labels via dual calibrative architecture, style confusion training, and label diversity regularization; 2) the evolved intermediate domain (EID) promotes the interaction between APL and DCG module, by mixing the labeled and unlabeled source domains in different manners. 
As for which kind of EID manner to choose, there is still some room for exploration, since the experimental results shows that both MixUp and ``X+U'' have their own advantages on different benchmarks. It indicates the significance to explore more effective EID manners for SSDG in the future.

\begin{table*}[t]  
\centering
\footnotesize
\resizebox{1.0\textwidth}{!}{
    \begin{tabular}{l|l|ccc|ccc|ccc|ccc|c}
    \toprule
    Method & Backbone & Ar$\!\scriptstyle{\rightarrow}\!$Cl & Ar$\!\scriptstyle{\rightarrow}\!$Pr & \multicolumn{1}{c|}{Ar$\!\scriptstyle{\rightarrow}\!$Rw} & Cl$\!\scriptstyle{\rightarrow}\!$Ar & Cl$\!\scriptstyle{\rightarrow}\!$Pr & \multicolumn{1}{c|}{Cl$\!\scriptstyle{\rightarrow}\!$Rw} & Pr$\!\scriptstyle{\rightarrow}\!$Ar & Pr$\!\scriptstyle{\rightarrow}\!$Cl & \multicolumn{1}{c|}{Pr$\!\scriptstyle{\rightarrow}\!$Rw} & Rw$\!\scriptstyle{\rightarrow}\!$Ar & Rw$\!\scriptstyle{\rightarrow}\!$Cl & \multicolumn{1}{c|}{Rw$\!\scriptstyle{\rightarrow}\!$Pr} & Avg. \\
    \midrule
    Source only~\cite{resnet} & \multirow{7}[2]{*}{ResNet-50} & 34.9  & 50.0  & 58.0  & 37.4  & 41.9  & 46.2  & 38.5  & 31.2  & 60.4  & 53.9  & 41.2  & 59.9  & 46.1  \\
    DAN~\cite{long2015learningDAN} &       & 43.6  & 57.0  & 67.9  & 45.8  & 56.5  & 60.4  & 44.0  & 43.6  & 67.7  & 63.1  & 51.5  & 74.3  & 56.3  \\
    DANN~\cite{dann} &       & 45.6  & 59.3  & 70.1  & 47.0  & 58.5  & 60.9  & 46.1  & 43.7  & 68.5  & 63.2  & 51.8  & 76.8  & 57.6  \\
    JAN~\cite{long2017JAN} &       & 45.9  & 61.2  & 68.9  & 50.4  & 59.7  & 61.0  & 45.8  & 43.4  & 70.3  & 63.9  & 52.4  & 76.8  & 58.3  \\
    CDAN~\cite{cdan} &       & 49.0  & 69.3  & 74.5  & 54.4  & 66.0  & 68.4  & 55.6  & 48.3  & 75.9  & 68.4  & 55.4  & 80.5  & 63.8  \\
    CDAN+E~\cite{cdan} &       & 50.7  & \textbf{70.6} & \textbf{76.0} & \textbf{57.6} & \textbf{70.0} & \textbf{70.0} & \textbf{57.4} & 50.9  & \textbf{77.3} & \textbf{70.9} & 56.7  & \textbf{81.6} & 65.8 \\
    \textbf{Ours w/ CDAN+E} &  & \textbf{60.9} & 72.4 & 74.3 & 56.9 & 73.8 & 69.3 & 53.5 & \textbf{57.6} & 75.5 & 65.9 & \textbf{66.7} & 83.9 & \textbf{67.6} \textcolor{red}{($\uparrow$1.8)} \\
    \midrule
    CDAN+E~\cite{cdan}  & \multirow{2}[2]{*}{ResNet-18} & 40.3  & 57.9  & 64.9  & 43.9  & 58.4  & 57.1  & 43.9  & 40.3  & 65.1  & 56.6  & 47.7  & 74.4  & 54.2  \\
    \textbf{Ours w/ CDAN+E} &       & 55.6 & 67.4 & 69.2 & 49.6 & 65.5 & 63.6 & 46.7 & 53.9 & 70.0 & 60.6 & 62.2 & 78.8 & 61.9 \textcolor{red}{($\uparrow$7.7)} \\
    \bottomrule
    \end{tabular} }
\vspace{-0.1cm}
\caption{Extension to unsupervised domain adaptation by comparing our method with other DA methods on \textbf{Office-Home} benchmark. ``Ar$\rightarrow$Cl'' means that Ar is the labeled source domain and Cl is the unlabeled target domain in DA task.}
\vspace{-0.3cm}
\label{tab:da_office}
\end{table*}

\subsection{Experiments on Open-Set SSDG}\label{sec:exp-open-set-ssdg}
We also conduct experiments on a more realistic and meaningful setting, namely open-set SSDG. The experiments are carried out on the newly-constructed web-crawled datasets, namely PACS-Webdata and Office-Webdata, where the experimental details refer to Section \ref{sec:exp-setup}.
Besides, due to the existence of web-labels in web data, we also develop different strategies to exploit these web-labels, which includes:
\begin{itemize}[leftmargin=12pt, topsep=2pt, itemsep=0pt]
    \item \textbf{Web-label ignored}: We ignore web-labels and simply treat web-crawled data as an unlabeled data, and perform DA/SSL/ours on both the labeled source domain data and unlabeled web-crawled data simultaneously, and finally, evaluate the model on the unseen target domain.
    \item \textbf{Web-label retained}: We treat web-labels as true label, and then train a DG/ours model on both the labeled source domain data and labeled web-crawled data simultaneously, and finally, evaluate the model on the unseen target domain. The whole process does not involve pseudo labeling.
    \item \textbf{Web-label refined} (exclusive to implement within our framework): We use web-labels to initialize pseudo labels that is refined by APL module from the second cycle, where APL module is not utilized in the first cycle. 
\end{itemize}

The training requirements of different web-label strategies are summarized in Table \ref{tab:opensetssdg}.  Our method is compared with other methods under different web-label strategies. In our framework, the APL module is implemented using either a close-set DA method, e.g. MCD~\cite{mcd}, which aims to classify the web-crawled data into $K$ classes, or an open-set DA method, e.g. OSDA~\cite{saito2018open}, which aims to categorize the web-crawled data into $K+1$ classes that include $K$ in-distribution classes and an out-of-distribution (OOD) class, where the OOD samples are rejected and only the in-distribution samples participate in the training of the DCG module. The setting of ``web-label retained'' does not require the use of the APL module, whereas both ``web-label ignored'' and ``web-label refined'' strategies are dependent on the APL module.
The comparison results of experiments on PACS-Webdata and Office-Webdata are reported in Table \ref{tab:pacs_web} and Table \ref{tab:office_web}, respectively. From these tables, we can see that:
\begin{itemize}
    \item Our method demonstrates consistent superior performance compared to other related methods in both SSL (MixMatch~\cite{MixMatch}, FixMatch~\cite{FixMatch}, MeanTeacher~\cite{Mean-teacher}) and DG (RSC~\cite{huang2020rsc}, MixStyle~\cite{mixstyle}, L2D~\cite{l2d}, DGvGS~\cite{DGvGs}) approaches on open-set SSDG tasks, as indicated by Table \ref{tab:pacs_web}. When in the ``web-label ignored'' strategy, our method surpasses SSL methods by disentangling SSDG into adaptation and generalization phases, which can avoid conflicting goals between the two phases, and the EID in our method ensures that the two phases mutually benefit each other. When in the ``web-label retained'' strategy, our method without APL still shows superiority to DG methods, since the DCG module can exploit both label and style information of web data, whereas other DG methods only focus on utilizing style information.
    \item Our method shows exceptional performance when using the ``web-label refined'' strategy, which outperforms the ``web-label retained" and ``web-label ignored" strategies on both PACS-Webdata and Office-Webdata benchmarks. It demonstrates the effectiveness of our method with APL that can refine web-labels to provide higher-quality pseudo labels. Therefore, an advanced DA method is expected to incorporate into APL module to further improve the performance on open-set SSDG.
    \item By comparing the results of Table \ref{tab:pacs} and Table \ref{tab:pacs_web}, as well as those of Table \ref{tab:office} and Table \ref{tab:office_web}, our method with open-set SSDG setting significantly outperforms to that with close-set SSDG setting, because the web-crawled data offers a wide variety of styles that the close-set dataset always lacks. This improvement confirms the positive impact of web-crawled data on SSDG, despite the out-of-distribution noise it may introduce. 
    We consider it worthwhile to explore better ways to utilize web-crawled data for SSDG and leave this as a future research topic.
\end{itemize}

\subsection{Extension to Unsupervised Domain Adaptation}
To stress the effectiveness of EID, our method is also plugged and played in the unsupervised domain adaptation tasks. The experiments are conducted on a popular DA benchmark, aka Office-Home dataset~\cite{office-home}. There are a total of 12 tasks in Office-Home. We conduct experiments on each task and report the average accuracy of all tasks as the final result.

\textbf{Setup}:
Unlike the SSDG task, the DA task requires training the model on both the labeled source domain and the unlabeled target domain. Following the standard DA protocols~\cite{cdan}, we train the APL and DCG modules on both the labeled source domain and unlabeled target domain. During evaluation, we discard the DCG module and only reserve the APL module to test its performance on the target domain.

\textbf{Implementation Details}:
We choose CDAN+E~\cite{cdan} as the baseline method to implement APL module using different backbones, i.e., ResNet-18 and ResNet-50, while keeping the backbone of DCG module unchanged (ResNet-18). 
The training hyper-parameters of APL are mostly inherited from its baseline~\cite{cdan}. And the hyper-parameters of DCG and EID can be referred to the aforementioned settings of SSDG. 

\textbf{Experimental Results and Analysis}:
The comparison of our method with multiple DA methods are reported in Table \ref{tab:da_office}. It shows that our method exhibits significant improvements in performance when compared to the baseline, particularly when using ResNet-18 as the APL backbone. This success can be attributed to that the DCG module acts as a strong teacher model to provide valuable knowledge to teach the APL module. In conclusion, our method demonstrates substantial success not only on SSDG task, but also on DA task, making it a promising area for further exploration.

\section{Conclusion}
In this paper, we propose a novel domain generalization paradigm, termed as \emph{Semi-Supervised Domain Generalization}, which aims to leverage the rich style information of unlabeled data to cooperate with limited labeled source data for domain generalization. To fully exploit the unlabeled data, we propose a framework equipped with \emph{Adversarial Pseudo Labeling} and \emph{Dual Calibrative Generalization} modules for label propagation and learning generalization, respectively. To encourage the modules to benefit mutually, we propose an \emph{Evolving Intermediate Domain} to bridge the labeled and unlabeled source domains, which can gradually narrow the domain discrepancy to improve the quality of pseudo labels of the APL module, thereby enhancing the generalization ability of the DCG module. Furthermore, we explore a more practical setting of SSDG, known as the open-set SSDG, by utilizing low-cost web-crawled data with distinct styles to facilitate domain generalization. To support this, we create two open-set SSDG datasets, namely PACS-Webdata and Office-Webdata, by crawling abundant web data.
We build extensive strong baselines to stress the effectiveness of our method both on the close-set and open-set SSDG settings. To sum up, our paper provides extensive strong baselines and abundant benchmarks for SSDG, which can promote the development of domain generalization community.

\ifCLASSOPTIONcaptionsoff
  \newpage
\fi

\bibliographystyle{IEEEtran}

\end{document}